%% file: main.tex
\documentclass[10pt,twocolumn,letterpaper]{article}

\usepackage{iccv}              


\definecolor{iccvblue}{rgb}{0.21,0.49,0.74}
\usepackage[pagebackref,breaklinks,colorlinks,allcolors=iccvblue]{hyperref}

\usepackage[accsupp]{axessibility}  

\def\paperID{P13N-6} 
\def\confName{ICCV}
\def\confYear{2025}

\title{Human Preference-Aligned Concept Customization Benchmark \\ via Decomposed Evaluation}

\usepackage{acronym}
\usepackage{amsmath}

\usepackage{amsfonts}
\usepackage{color}
\usepackage{bbm}
\usepackage{bm}
\usepackage{mathcomp}
\usepackage{comment}
\usepackage{multirow}
\usepackage{graphicx}
\definecolor{mygray}{gray}{0.937254902}
\usepackage{booktabs} 
\usepackage{longtable}
\usepackage{float}
\usepackage{amssymb}
\usepackage{pifont}
\newcommand{\cmark}{\ding{51}}%
\newcommand{\xmark}{\ding{55}}%
\acrodef{LLM}[LLM]{Large Language Model}
\acrodef{MLLM}[MLLM]{Multimodal Large Language Model}
\acrodef{ours}[CC-AlignBench]{Human Preference-Aligned Concept Customization Benchmark}
\acrodef{our_eval}[D-GPTScore]{Decomposed GPT Score}
\acrodef{XGBoost}[XGBoost]{eXtreme Gradient Boosting}
\acrodef{VLM}[VLM]{Vision-Language Model}
\acrodef{LMM}[LMM]{Large Multimodal Models}
\acrodef{VQA}[VQA]{Visual Question Answering}
\acrodef{ViT}[ViT]{Vision Transformer}
\acrodef{MMD}[MMD]{Maximum Mean Discrepancy}

\author{
    Reina Ishikawa\textsuperscript{1},
    Ryo Fujii\textsuperscript{1},
    Hideo Saito\textsuperscript{1},
    Ryo Hachiuma\textsuperscript{2}
    \\
\textsuperscript{1}Keio University, \textsuperscript{2}NVIDIA\\
{\tt\small \{reina.ishikawa, ryo.fujii0112, hs\}@keio.jp, rhachiuma@nvidia.com}
}

\begin{document}

\newcommand{\update}[1]{{\textcolor{blue}{#1}}}
\newcommand{\insertTable}{\textcolor{red}{---InsertTable---}}
\newcommand{\insertFigure}{\textcolor{red}{---InsertFigure---}}
\newcommand{\citeme}{\textcolor{blue}{<citation>}}
\newcommand{\tb}{Table~}
\newcommand{\fig}{Figure~}
\newcommand{\sect}{Section~}
\newcommand{\eq}{Equation~}
\newcommand{\apx}{Appendix~}

\newcommand{\writeme}{writeme writeme writeme writeme writeme writeme writeme writeme writeme writeme writeme writeme writeme writeme writeme writeme writeme writeme writeme writeme writeme writeme writeme writeme writeme writeme writeme writeme writeme writeme writeme writeme writeme.}

\maketitle 
\input{latex/01_abstract}

\begin{figure}[tbh]
  \centering
  \includegraphics[width=\linewidth]{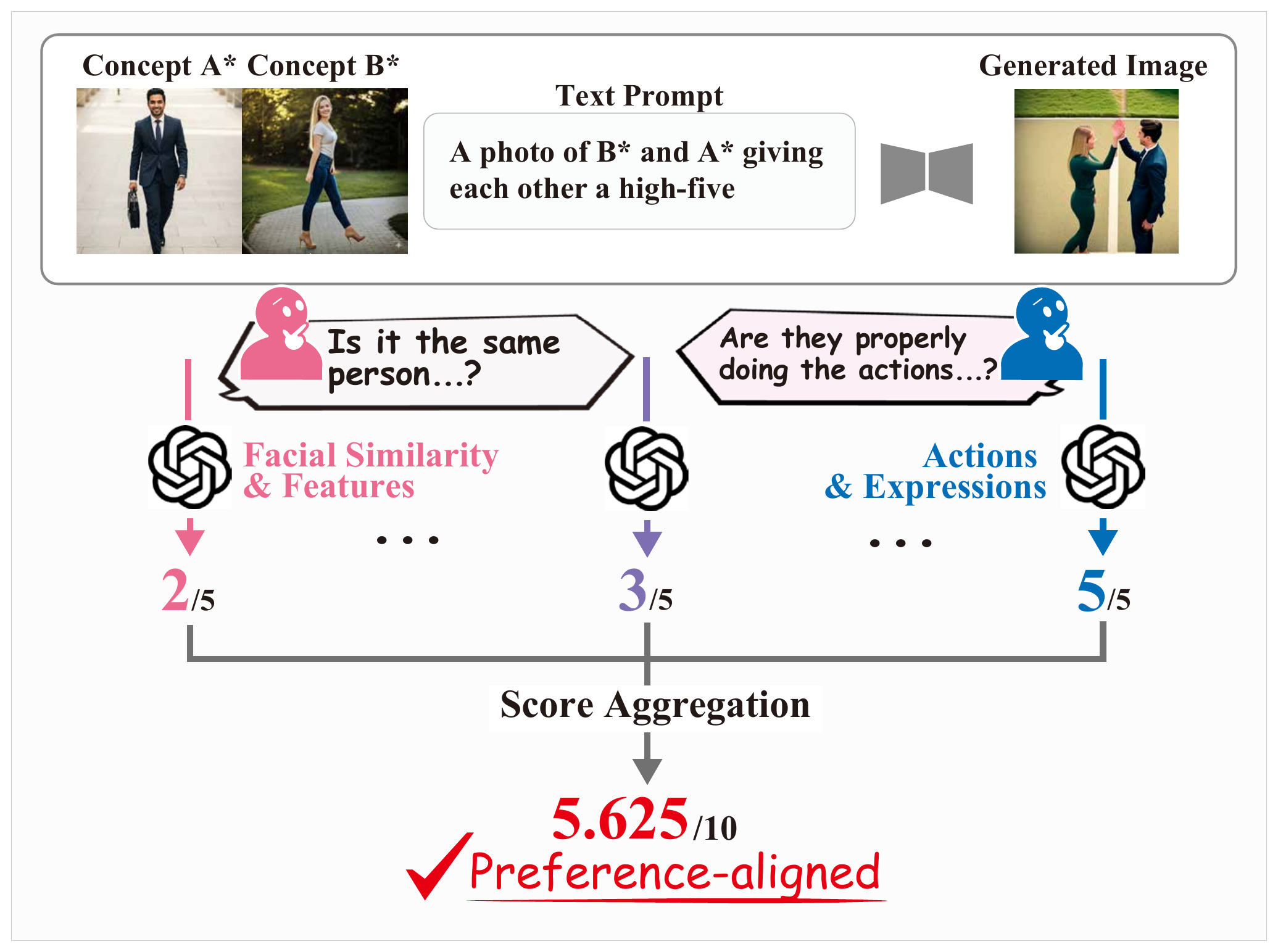}
  \caption{
  \textbf{\ac{our_eval}} evaluates images generated by concept customization through a two-step process: aspect-wise evaluation followed by aggregation. This approach achieves significantly higher correlation with human preference scores than existing methods, establishing a more reliable and human-aligned metric.
  }
  \label{fig:teaser}
\end{figure}

\begin{figure*}[t]
  \centering
  \includegraphics[width=0.9\linewidth]{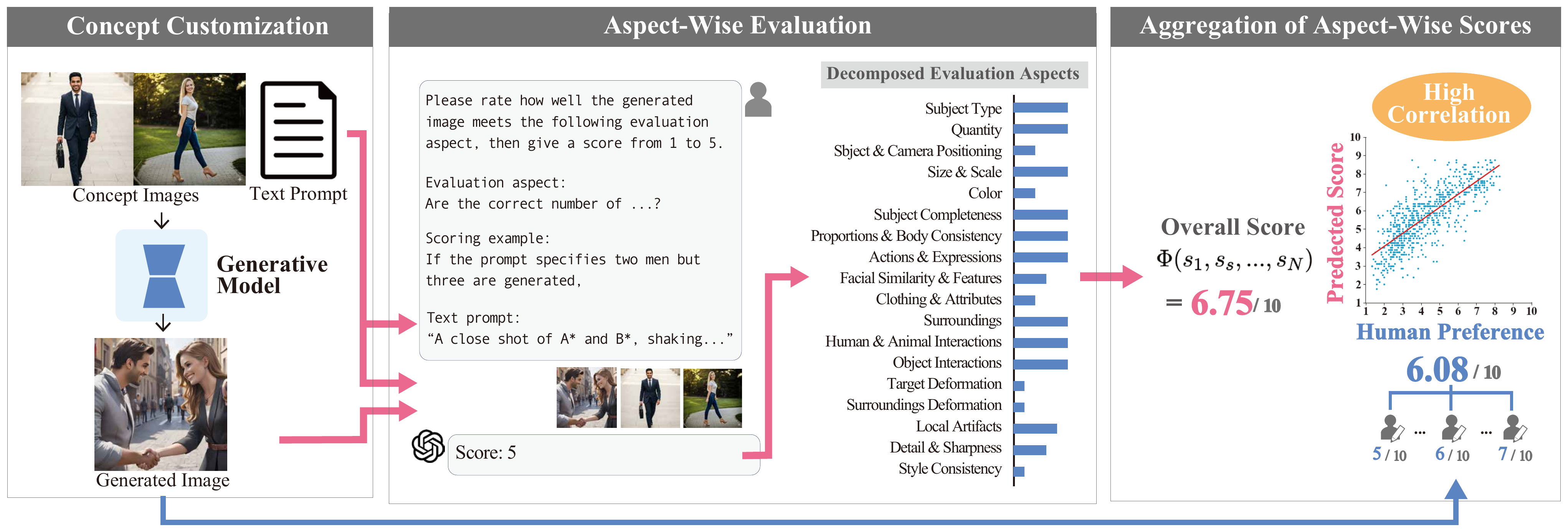}
  \caption{
  The pipeline of \ac{our_eval}. This metric comprises two phases: (1) \textbf{Aspect-wise evaluation with \ac{MLLM}}, and (2) \textbf{Aggregation of aspect-wise scores}. }
  \label{fig:flow}
\end{figure*}

\section{Introduction}
\label{sec: intro}
\input{latex/02_introduction}  
\section{Related Work}
\label{sec: relatedwork}
\input{latex/03_relatedwork} 

\begin{table*}[t]
\centering
\caption{\textbf{Composition of the \ac{ours} dataset}, divided into three difficulty levels based on the number of persons and interaction types.}
  \label{tab: dataset_composition}
\small
\begin{tabular}{lccccl}
\hline
\textbf{Difficulty} & \textbf{\#Total} & \textbf{\#Base} & \textbf{\#Persons per prompt} & \textbf{\#Different Actions}& \textbf{Action Type} \\ \hline
easy   & 260 & 52  & 1 & 13 & non-interaction\\
medium & 260 & 52  & 1 & 13 & non-interaction\\
hard   & 460 & 92  & 2 & 23 & mutual interaction\\ \hline
\end{tabular}
\end{table*}

\input{latex/04_method}  
\section{Experiments}
\label{sec: experiments}
\input{latex/05_experiments}  
\section{Discussion}
\label{sec: discussion}
\input{latex/06_discussion}  

\section{Limitations}
\label{sec: limitation}
\input{latex/07_limitation}

\section{Ethical Considerations}
\label{sec: ethicstatement}

In creating our benchmark dataset, we carefully removed harmful, offensive, or abusive expressions. Since the dataset is intended solely for image evaluation, it poses no direct social risks. However, it is important to note that terms such as ``hit'' and ``kick'' are included to introduce variations in mutual interactions.  As generated images heavily depend on the training data, controlling the output is challenging. To respect portrait rights, only AI-generated reference images are used, eliminating concerns about harm to individuals. Additionally, in our experiments, visual inspections were conducted to ensure the images do not contain harmful or offensive content. Nevertheless, toxic images may still be generated depending on the model. Users should be aware of the potential risks and refrain from using this dataset to generate offensive or inappropriate images, especially of real individuals. Caution is advised when releasing these images.

\section{Conclusion}
\input{latex/08_conclusion}  

\section*{Acknowledgment}
This work was supported by JST BOOST, Japan Grant Number JPMJBS2409.

{
    \small
    \bibliographystyle{ieeenat_fullname}
    \bibliography{main}
}

\clearpage 

\input{latex/99_appendix}

\end{document}

%% file: latex/01_abstract.tex
\begin{abstract}
Evaluating concept customization is challenging, as it requires a comprehensive assessment of fidelity to generative prompts and concept images.
Moreover, evaluating multiple concepts is considerably more difficult than evaluating a single concept, as it demands detailed assessment not only for each individual concept but also for the interactions among concepts.
While humans can intuitively assess generated images, existing metrics often provide either overly narrow or overly generalized evaluations, resulting in misalignment with human preference.  
To address this, we propose \textbf{\ac{our_eval}}, a novel human-aligned evaluation method that decomposes evaluation criteria into finer aspects and incorporates aspect-wise assessments using \ac{MLLM}.  
Additionally, we release \textbf{\ac{ours}}, a benchmark dataset containing both single- and multi-concept tasks, enabling stage-wise evaluation across a wide difficulty range—from individual actions to multi-person interactions.  
Our method significantly outperforms existing approaches on this benchmark, exhibiting higher correlation with human preferences.  
This work establishes a new standard for evaluating concept customization and highlights key challenges for future research. The benchmark and associated materials are available at \href{https://github.com/ReinaIshikawa/D-GPTScore}{https://github.com/ReinaIshikawa/D-GPTScore}
\end{abstract}

%% file: latex/02_introduction.tex
\textit{When evaluating AI-generated images from user-provided text prompts, what criteria do you employ to determine whether the generated images are of high quality?}
You may implicitly consider diverse evaluation aspects such as overall realism and fidelity to the text prompt.
Although humans can intuitively and comprehensively evaluate generated images based on these implicit aspects, automatic methods that achieve comprehensive evaluation aligned with human preferences remain underexplored in the research community.

In recent years, the rise of diffusion models~\cite{Ho2020Denoising, Sohl-Dickstein2015Deep, rombach2022highresolution, podell2023sdxl} has led to numerous text-to-image generation methods.
Building upon these approaches, concept customization~\cite{voynov2023p+, ruiz2023dreambooth, Zhang2023ControleNet} aims to extend pretrained diffusion models to support single or multiple personalized concepts using only a few reference images per concept, combined with user-provided text prompts that condition the concepts (e.g., \texttt{A$^\ast$ is standing}, where \texttt{A$^\ast$} denotes the concept).

To advance controllable image generation—hereafter referred to as \textit{concept customization}—rigorous evaluation of generated images is essential to appropriately guide research progress.
However, due to several critical issues in existing evaluation metrics and datasets, standardized benchmarks for the concept customization task remain underdeveloped.

First, \textbf{existing evaluation metrics are not aligned with human preferences}.
Humans can comprehensively consider various evaluation criteria and implicitly weigh important aspects to form an overall judgment of generated image quality. In contrast, existing evaluation metrics for concept customization~\cite{Hinz2022SOAIscore, Phung2024AttentionRefocusing, Dengw2022arcface} either assess only partial aspects or provide overly abstracted evaluations of generated images, resulting in low correlation with human preference scores (see \sect\ref{subsec: correlation} for experimental results).

Second, current benchmarks~\cite{Kumari2023customdiffusion} only partially evaluate concept customization models, and \textbf{the evaluation settings are limited to relatively simple cases}.
Most studies employ text-to-image benchmarks~\cite{saharia2022photorealistic, Bakr_2023_HRS-Bench, Hu2023TIFA, gokhale2022benchmarking} for concept customization tasks; however, these datasets contain incomplete inputs and evaluate only partial aspects of the task.
Although CustomConcept101~\cite{Kumari2023customdiffusion} is a widely used and comprehensive benchmark, the target concepts and text prompts are relatively simple for customization models.
For example, the target concepts predominantly consist of objects that do not interact with scenes or other concepts; the reference images are close-up shots of the concepts; and the text prompts for concepts describe non-interactive scenarios (\eg, \texttt{A$^\ast$ is walking}).

In this paper, we propose \ac{ours}, which includes a human-aligned evaluation metric and a comprehensive evaluation dataset for the concept customization task. Inspired by LLM-as-a-Judge~\cite{Lianmin2023NeurIPS}, we leverage the capabilities of state-of-the-art \acp{MLLM}, particularly GPT-4o~\cite{openai2024gpt4technicalreport}, to develop a human-aligned evaluation metric. These \acp{MLLM} demonstrate excellent multi-modal language understanding by processing combined text and visual inputs, enabling tasks such as image captioning and visual question answering. This multi-modal understanding makes them well-suited for evaluating concept customization, which requires consideration of both text prompts and reference images.

However, if GPT-4o directly assesses the entire generated image at once, it may only provide a coarse overall score. To address this, we propose to \textit{decompose} the evaluation criteria into multiple predefined aspects and evaluate each aspect individually, thereby preventing the omission of detailed factors and ensuring better alignment with human preferences. The individual aspect scores are then aggregated to produce the final evaluation metric.

Furthermore, we construct a dataset consisting of pairs of text prompts and reference images, focusing specifically on humans exhibiting more complex and diverse interactions with scenes and/or other humans. The dataset contains 980 text prompts encompassing three levels of human actions (\ie, a single person's action, two persons' independent actions, and two persons' mutual actions), and five conditioning types (\ie, five different combinations of action, layout, expressions, and surroundings) (see \sect\ref{subsec: dataset} for details). By evaluating according to the levels of human actions or conditioning types, this dataset enables a systematic assessment of a model's concept customization capabilities.

Our main contributions can be summarized as follows:
\begin{itemize}
\item We propose the \textbf{\ac{ours}} for comprehensive evaluation of both single- and multi-concept customization, offering varying difficulty levels based on character count and scene complexity, thus allowing gradual assessment of model capabilities.
\item We introduce an evaluation metric, \textbf{\ac{our_eval}}, leveraging \ac{MLLM} that enables comprehensive and automated evaluation of concept customization. Our metric demonstrates better alignment with human preferences than existing methods and supports multi-concept evaluation.
\item We conduct extensive ablation studies using the \ac{ours} dataset to verify the effectiveness of our proposed metric. Experimental results confirm that our metric correlates more strongly with human preference scores than existing evaluation metrics.
\end{itemize}

We believe this benchmark will advance the concept customization research community toward more realistic and human preference-aligned evaluation.

%% file: latex/03_relatedwork.tex
\subsection{Metrics for concept customization}
\label{subsec: metric_for_concept_customization}
Evaluating the quality of generated images using appropriate metrics and obtaining feedback is crucial for model improvement. However, existing methods fail to comprehensively assess all relevant aspects of concept customization, resulting in evaluations that diverge from human preferences.

Metrics such as Counting~\cite{Phung2024AttentionRefocusing}, SOA-I Score~\cite{Hinz2022SOAIscore}, Compositions~\cite{Phung2024AttentionRefocusing}, and YOLO Score~\cite{Li2021yoloscore} can recognize the number of subjects and layout in multi-subject scenarios, but their assessments are limited and do not evaluate verbs in the text prompt (\eg, actions). Conversely, methods like the CLIP series~\cite{Radford2021CLIP} and DINO Score~\cite{Caron2021Dino} provide overall evaluations by comparing images with text or texts with each other; however, their assessments are general and insufficient to capture fine-grained differences. QS~\cite{Gu2020QS}, CLIP Aesthetic~\cite{schuhmann2022laion5bopenlargescaledataset}, and IS~\cite{Gu2020QS} demonstrate high correlation with human ratings but are restricted to assessing image quality. 
ImageReward~\cite{xu2023imagereward} and VQA Score~\cite{Lin2024VQAScore} enable integrated evaluation of a text prompt and a generated image through a preference-based model and Question Answering, respectively. However, both models are primarily designed for text-to-image tasks.
Although GPT-4V Score~\cite{zhong2024multilora} can evaluate both composition quality and image quality using \ac{MLLM}, it does not account for multi-concept scenarios.
In contrast, our proposed method enables a comprehensive evaluation of generated images across all necessary aspects, including fidelity to mutual interactions.

\subsection{Benchmarks for concept customization}
\label{subsec: benchmarks_for_concept}

Proper evaluation of images generated through concept customization requires the consideration of both text prompts and reference images alongside the outputs. However, many existing datasets lack either prompts or reference images~\cite{zhou2021simple, saharia2022photorealistic, gokhale2022benchmarking, Bakr_2023_HRS-Bench, ruiz2023dreambooth, Hu2023TIFA}. Even when both are available, reference images often depict multiple individuals~\cite{LinCOCO2014, Plummer2015Flickr30k, young2014From, Schuhmann2021LAION400MOD, cho2023DALL-EVAL} or focus solely on faces or upper bodies~\cite{Karras2019Flickr-Faces-HQ, Kumari2023customdiffusion, peng2024dreambench}, limiting their suitability for representing individual concepts. 
ImagenHub~\cite{ku2024imagenhub} is composed of pairs of concept images and corresponding text prompts, but excludes human subjects.
As a result, no existing dataset adequately supports the evaluation of dynamic movements in concept customization.
Accordingly, current methods often rely on combinations of such datasets and their associated metrics. Furthermore, datasets featuring complex actions, such as mutual interactions, remain scarce, and no benchmark explicitly addresses them.

To bridge these gaps, we introduce a dataset that includes both prompts and reference images for single- and multi-concept scenarios. It also provides prompts of varying difficulty, enabling step-wise evaluation of generation performance.

\subsection{MLLM-based human-preference scoring}
\label{subsec: llmbased_human_preference_scoring}
Recent research has actively explored scoring methods aligned with human preferences using \acp{MLLM}, particularly for video evaluation~\cite{bansal2024videophy, Bansal2025videophy2, Wang2024AIGV-Assessor, Nema2018towards}. To improve alignment, several methods assess generative models by defining various aspects and scoring each individually~\cite{wu2024qalign, He2024VideoScore}. Some works aggregate aspect-wise scores based on human preferences, such as MMHE~\cite{Ohi2024MultimodalMM}, which employs harmonic weighting, and MetaMetrics~\cite{winata2025metametrics}, which trains a regression model to combine the outputs of existing metrics. 

For MLLM-based metrics in concept customization tasks, VIEScore~\cite{ku2024viescore} and CIGEval~\cite{wang2025cigeval} proposed evaluation metrics by decomposing the task, but neither accounts for the action fidelity of interactions between living subjects.
DreamBench++~\cite{peng2024dreambench} improves alignment with human scores on single-concept customization tasks by performing decomposition and outputting the reasoning process for task understanding. However, it does not explicitly generate scores for each evaluation aspect, which may reduce the reliability of the final score due to implicit reasoning~\cite{yu2025llmsreallythinkstepbystep}. 

Our method is unique in that it achieves human-aligned evaluation for concept customization---supporting both single- and multi-concept tasks---by performing unified explicit aspect-wise evaluation and integration using a single \ac{MLLM}.

%% file: latex/04_method.tex
\section{D-GPTScore Metric}
\label{sec: method}

Our concept customization evaluation metric is straightforward and consists of two phases: (1) \textbf{Aspect-wise evaluation using \ac{MLLM}}, and (2) \textbf{Aggregation of aspect-wise scores}. As a preliminary step, we provide a brief overview of concept customization as follows:

    \noindent \textbf{Concept Customization.} An image $I_g$ is generated from a text prompt $\mathcal{T} = \{t_1, t_2, ..., t_k\}$, which comprises several short text elements concatenated with commas, and a set of reference images $\mathcal{I} = \{i_1, i_2, ..., i_l\}$, using the generative model $\theta$:
\begin{equation}
  \label{eq: image_generation}
  I_g = \mathcal{\theta}(\mathcal{T}, \mathcal{I}).
\end{equation}

\subsection{Aspect-wise evaluation with MLLM}
\label{subsec: aspect_wise_eval_with_mllm}

In the aspect-wise evaluation phase, we first decompose the evaluation criteria for concept customization into $N$ predefined aspects $\mathcal{A} = \{a_1, a_2, ..., a_{N}\}$, considering not only single- but also multi-concept scenarios (see \sect\ref{subsec: eval_aspects} for details). To obtain the $n$-th aspect-wise score $s_n \in \mathbb{R}$ of the generated image $I_g$, we independently feed the $n$-th evaluation aspect $a_n$ into a \ac{MLLM} $\phi$, along with the text prompt $\mathcal{T}$ and reference images $\mathcal{I}$:
\begin{equation}
\label{eq: aspectwise_score_re}
s_n = \phi(a_n, I_g, \mathcal{T}, \mathcal{I}),
\end{equation}

\subsection{Aggregation of aspect-wise scores}
\label{subsec: aggregation_of_aspectwise_scores}

Once the scores ${s_1, s_2, ..., s_N}$ for multiple aspects are obtained, the aggregation model $\Phi$ is applied to compute the overall score $s_{overall}$ on a scale of $[1,10]$ as follows:
\begin{equation}
\label{eq: aggregation}
s_{overall} = \Phi(s_1, s_2, ..., s_N).
\end{equation}
An ablation study on the choice of aggregation model is provided in \sect\ref{subsec: ablation}.

\subsection{Identification of evaluation aspects}
\label{subsec: eval_aspects}

We predefine the evaluation aspects into two main categories: (1) Concept Fidelity and (2) Quality Assessment. 

``Concept Fidelity'' assesses the adherence of the generated image to the text prompt and reference images and is further divided into the following four subcategories, totaling 13 aspects. 

\noindent \textbf{1) Object Existence\&Accuracy} assesses whether the specified objects or people are generated according to the prompt, in appropriate quantities, based on the \textit{Subject Type} and \textit{Quantity}.

\noindent \textbf{2) Layout \& Composition Fidelity} assesses whether the \textit{Subject \& Camera Positioning} (\ie, relative positions, including camera angles) and the relative \textit{Size \& Scale} of the generated characters and objects align with the instructions in the prompt.

\noindent \textbf{3) Object-Level Fidelity} assesses whether the generated objects, people, and surroundings are accurately depicted, maintaining consistency with both the text and reference images in terms of \textit{Color}, \textit{Proportions \& Body Consistency}, \textit{Actions \& Expressions}, \textit{Facial Similarity \& Features}, \textit{Clothing \& Attributes}, and \textit{Surroundings}.

\noindent \textbf{4) Multi-Concept Interaction Consistency} assesses whether interactions between humans (\ie, \textit{Human \& Animal Interactions}) or between objects (\ie, \textit{Object Interactions}) are generated in a natural and coherent manner, as specified in the text prompt.

``Quality Assessment'' evaluates image quality, consisting of 5 aspects: \textit{Subject Deformation}, \textit{Surroundings Deformation}, \textit{Local Artifacts}, \textit{Detail\&Sharpness}, and \textit{Style Consistency}.

\begin{table*}[]
\centering
\caption{\textbf{Example text prompt in \ac{ours}}. Each prompt consists of the four base prompts: action, layout, expression, and surroundings.}
\label{tab: example_text_prompt}
\small
\begin{tabular}{ll}
\hline
\textbf{type} & \textbf{Example Text Prompt}
\\\hline\hline
\begin{tabular}[c]{@{}l@{}}
action-only\end{tabular}
&\begin{tabular}{p{11.5cm}}
A photo of a woman putting her arm around a man's shoulder, Ultra HD quality. 
\end{tabular}
\\\hline

\begin{tabular}[c]{@{}l@{}}
action\\+layout\end{tabular}
&\begin{tabular}{p{13cm}}
A high angle shot of a woman putting her arm around a man's shoulder, standing close, Ultra HD quality. 
\end{tabular}
\\\hline

\begin{tabular}[c]{@{}l@{}}
action\\+expression\end{tabular}
&\begin{tabular}{p{13cm}}
A photo of a woman putting her arm around a man's shoulder, both looking amused, Ultra HD quality.
\end{tabular}
\\\hline

\begin{tabular}[c]{@{}l@{}}
action\\+surroundings\end{tabular}
&\begin{tabular}{p{13cm}}
A photo of a woman putting her arm around a man's shoulder, in an open green park, Ultra HD quality.
\end{tabular}
\\\hline

\begin{tabular}[c]{@{}l@{}}
all\end{tabular}
&\begin{tabular}{p{13cm}}
 A high angle shot of a woman putting her arm around a man's shoulder, standing close, both looking amused, in an open green park, Ultra HD quality. 
\end{tabular}
\\\hline
\end{tabular}
\end{table*}

\section{Dataset}
\label{subsec: dataset}

\subsection{Text preparation}
\label{subsec: text_preparation}

As shown in \tb\ref{tab: dataset_composition}, \ac{ours} consists of 196 prompts divided into three difficulty levels: 52 \textit{Easy}, 52 \textit{Medium}, and 92 \textit{Hard} prompts (referred to as \textbf{difficulty levels}).
\textit{Easy} prompts describe a single person acting; \textit{Medium} prompts involve two individuals performing independent actions such as ``standing” or ``walking''; and \textit{Hard} prompts depict interactions between two individuals, such as ``hugging'' or ``whispering.''

Each of the 196 base prompts comprises four elements: \textit{action, layout, expression}, and \textit{surroundings}.
To evaluate the model's expressive capabilities, five variations are generated from each base prompt: \textit{action-only} prompt ${t_{act}}$, \textit{action+layout} prompt ${t_{act}, t_{lt}}$, \textit{action+expression} prompt ${t_{act}, t_{exp}}$, \textit{action+surroundings} prompt ${t_{act}, t_{surr}}$, and \textit{all} prompt ${t_{act}, t_{lt}, t_{exp}, t_{surr}}$. Since each of the 196 prompts is expanded into these five variations, the total number of prompt variations in \ac{ours} amounts to $196 \times 5 = 980$.
An example text prompt is shown in \tb\ref{tab: example_text_prompt}.

\subsection{Image preparation}
\label{subsec: text_preparation}

To address concerns related to portrait rights, we used generative AI to create images without relying on photographs of real individuals.
First, we generated one image each of a man and a woman using \textit{Gemini 2.5 Flash Preview}~\cite{geminiteam2025geminifamilyhighlycapable}. Then, using \textit{Gemini 2.0 Flash Preview Image Generation}, we instructed the model to produce various patterns (\eg, close-up, long shot, frontal, and profile views) of the same man and woman.
We added 19 additional images per person, resulting in 20 images per individual and a total of 40 images. For models that require a single input image, we ensured the same full-body image was consistently used.

%% file: latex/05_experiments.tex
\begin{table*}[h]
\centering
\caption{\textbf{Comparison of correlation with human preference on \ac{ours}}. The left shows Pearson’s correlation coefficient, and the right shows Spearman’s rank correlation (higher is better). A correlation of $0.5$ or higher is generally considered strong, and the proposed metric significantly exceeds this threshold for the overall score and most individual models.
}
\label{tab: correlation}
\resizebox{\linewidth}{!}{
\begin{tabular}{l|cccccc|c}
\hline
Metric 
& \multicolumn{1}{c}{\begin{tabular}[c]{@{}c@{}}CustomDiffusion\end{tabular}}
& \multicolumn{1}{c}{\begin{tabular}[c]{@{}c@{}}OMG\\+LoRA\end{tabular}} 
& \multicolumn{1}{c}{\begin{tabular}[c]{@{}c@{}}OMG\\+InstantID\end{tabular}} 
& \multicolumn{1}{c}{\begin{tabular}[c]{@{}c@{}}FastComposer\end{tabular}} 
& \multicolumn{1}{c}{\begin{tabular}[c]{@{}c@{}}Mix-of-Show\end{tabular}}  
& \multicolumn{1}{c|}{\begin{tabular}[c]{@{}c@{}}DreamBooth\end{tabular}} 
& \multicolumn{1}{c}{\begin{tabular}[c]{@{}c@{}}Overall\end{tabular}}  \\ \hline \hline
ArcFace & 0.34 / 0.26 & 0.05 / 0.01 & -0.01 / 0.06 & 0.21 / 0.27 & 0.50 / 0.33 & 0.21 / 0.15 & 0.23 / 0.04 \\
CLIP T2I & 0.20 / 0.27 & -0.01 / 0.13 & 0.11 / 0.08 & 0.09 / 0.36 & 0.21 / 0.38 & 0.20 / 0.33 & 0.29 / 0.42 \\
CLIP T2T & 0.01 / 0.27 & -0.04 / -0.07 & -0.04 / -0.09 & 0.16 / 0.20 & 0.19 / 0.23 & 0.16 / 0.21 & 0.14 / 0.21 \\
CLIP Aes. & 0.46 / 0.09 & 0.14 / 0.04 & 0.20 / 0.00 & 0.26 / 0.15 & 0.29 / 0.02 & 0.44 / 0.26 & 0.51 / 0.49 \\
DINO & 0.28 / 0.10 & 0.04 / 0.14 & 0.30 / 0.10 & -0.17 / 0.09 & -0.07 / -0.20 & 0.26 / -0.02 & 0.10 / 0.04 \\ \hline
Ours & \textbf{0.80} / \textbf{0.54} & \textbf{0.66} / \textbf{0.34} & \textbf{0.70} / \textbf{0.47} & \textbf{0.64} / \textbf{0.46} & \textbf{0.65} / \textbf{0.44} & \textbf{0.64} / \textbf{0.38} & \textbf{0.78} / \textbf{0.69} \\ \hline
\end{tabular}
}
\end{table*}

\subsection{Experimental setup}
\paragraph{Baselines of evaluation metrics} 
To verify whether the proposed method aligns with human preferences, we compared it with five existing evaluation metrics commonly used for concept customization: identity preservation score by ArcFace~\cite{Dengw2022arcface}, CLIP T2T score~\cite{Radford2021CLIP}, CLIP T2I score~\cite{Radford2021CLIP}, CLIP Aesthetic Score~\cite{schuhmann2022laion5bopenlargescaledataset}, and DINO score~\cite{Caron2021Dino}. See our supplementary material for the details.

\paragraph{Target generative models} 
We evaluated five widely used and state-of-the-art image generation models that support multi-concept customization: CustomDiffusion~\cite{Kumari2023customdiffusion}, OMG~\cite{Kong2024OMG} with LoRA or InstantID, FastComposer~\cite{Xiao2024FastComposer}, Mix-of-Show~\cite{Gu2023Mix-of-show}, and DreamBooth~\cite{ruiz2023dreambooth}, using our proposed benchmark.

\paragraph{Aggregation model selection} 
For the aggregation model $\Phi$, we adopted the average function:
\begin{equation}
\label{eq: average}
\Phi(s_1, s_2, ..., s_N) = \frac{1}{N} \sum_{n=1}^{N} s_n,
\end{equation}
meaning that each aspect is treated equally to compute the final score. The choice of aggregation model is further discussed in \sect\ref{subsec: ablation}.

\paragraph{MLLM setting} We employed \texttt{gpt-4o-2024-08- 06}~\cite{openai2024gpt4technicalreport} as the MLLM for aspect-wise evaluation. All generated and reference images fed to GPT-4o were resized to 512$\times$512 pixels. For evaluation aspects where only generated images are provided (\eg, deformations or local artifacts), resolution can affect accuracy.
Therefore, we additionally provided GPT with two square-cropped images, each covering $50\%$ of the original size, centered vertically and placed on the left and right halves, along with the original image.

\paragraph{Preference score annotation} 
To evaluate human alignment, we measured the correlation between predicted scores and human preference scores. The human scores were obtained from 12 expert annotators (5 females, 7 males), all members of our laboratory.
We randomly selected 40 prompts from each of the three difficulty levels (\ie, \textit{easy}, \textit{medium}, \textit{hard}) and generated images using six different models for each prompt.
The resulting 720 images were evaluated by the annotators, who scored them from 1 to 10 with reference to both the text prompts and the reference images.
We then averaged the scores across annotators to obtain a single human preference score per image. The annotation took approximately two hours per annotator. Each annotator was compensated \$48 according to the current exchange rate.

\paragraph{Experimental environment} 
All experiments were conducted on a system equipped with a single NVIDIA RTX A6000 (48GB), an Intel(R) Core(TM) i9-9940X CPU, and the OpenAI API.

\begin{figure}[tb]
    \centering
        \includegraphics[width=\linewidth]{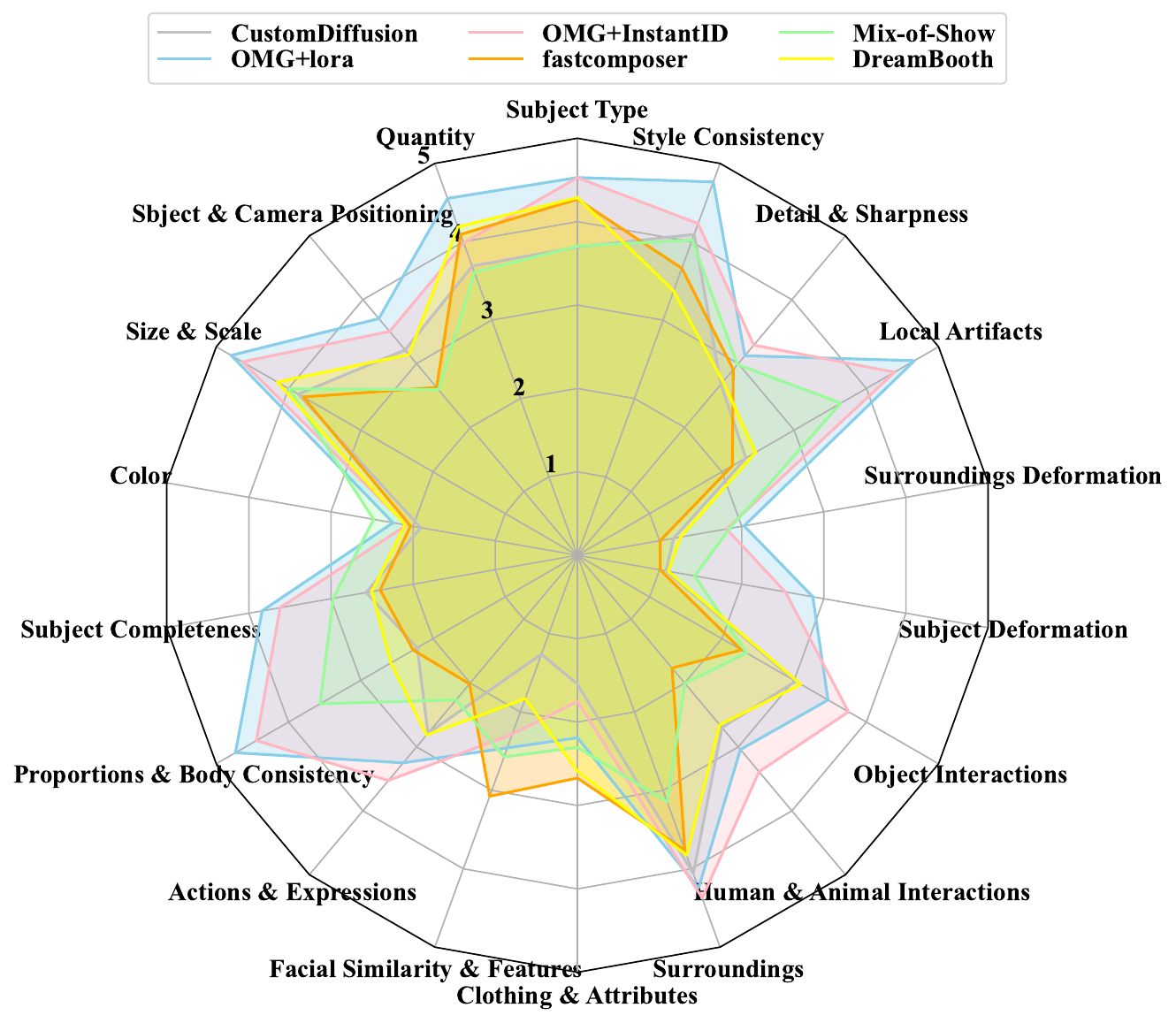}
        \caption{\textbf{Benchmark results of aspect-wise evaluation}. Each aspect is scored on a scale from 1 to 5, with higher values indicating better performance.}
        \label{fig: benchmark_raderchart}
\end{figure}

\subsection{Correlation with human preference}
\label{subsec: correlation}

For each evaluation metric, we computed the Pearson’s correlation coefficient and Spearman’s rank correlation between the human preference scores and the predicted scores.
The results are summarized in \tb\ref{tab: correlation}.
Our metric achieves an overall Pearson’s correlation of $0.78$ and an overall Spearman’s rank correlation of $0.69$ with human preferences, significantly surpassing all existing metrics. 
These results demonstrate that the proposed method provides human-aligned evaluation for concept customization and outperforms existing metrics. Furthermore, since omission of any key aspect may reduce correlation, the strong correlation exceeding 0.7 suggests that the proposed decomposed aspects are necessary and sufficient.

\subsection{Benchmark results}
\label{subsec: benchmark}


\begin{table*}[]
\centering
\caption{\textbf{Average benchmark scores for each difficulty level on \ac{ours}, predicted using \ac{our_eval}}. Each score ranges from 1 to 10, with higher values indicating better image generation performance. When applying linear regression, the leave-one-out approach is employed to ensure that the generated images from the model under evaluation are excluded from the training data.}
\label{tb: benchmark_difficulty_level}
\resizebox{0.97\textwidth}{!}{
\begin{tabular}{l|cccccc}
\hline
\textbf{Difficulty Level} &
  \multicolumn{1}{c}{\textbf{CustomDiffusion}} &
  \multicolumn{1}{c}{\textbf{OMG+LoRA}} &
  \multicolumn{1}{c}{\textbf{OMG+InstantID}} &
  \multicolumn{1}{c}{\textbf{FastComposer}} &
  \multicolumn{1}{c}{\textbf{Mix-of-Show}} &
  \multicolumn{1}{c}{\textbf{DreamBooth}} \\ \hline\hline
Easy   & 5.64 & \textbf{7.28} & 7.13 & 5.21 & 5.75 & 5.53 \\
Medium  & 4.62 & \textbf{7.05} & 6.64 & 4.72 & 5.27 & 5.01 \\ 
Hard    & 4.30 & \textbf{6.41} & 6.10 & 4.45 & 4.82 & 4.87 \\ \hline
Overall  & 4.74 & \textbf{6.81} & 6.52 & 4.72 & 5.19 & 5.09 \\  \hline
\end{tabular}
}
\end{table*}


\begin{table*}[t]
\centering
\caption{\textbf{\ac{our_eval} scoring examples on the \textit{Hard} level subset of \ac{ours}}.
For each model, the top three scores are highlighted in red, blue, and yellow, respectively, to illustrate their correspondence with the human preference rankings.}
\label{tb: scoring_examples}
  \includegraphics[width=0.97\textwidth]{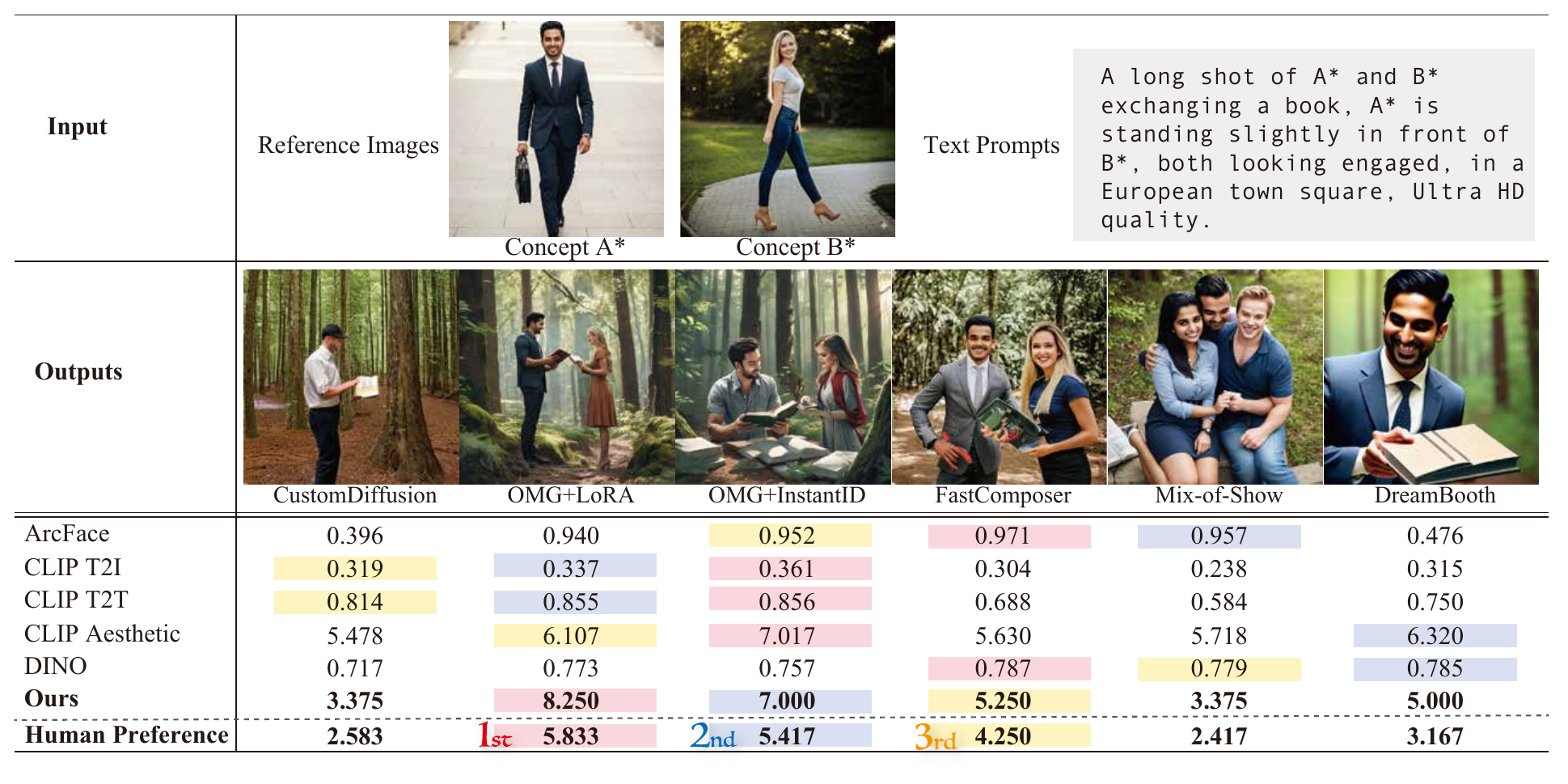}
\end{table*}

\tb\ref{tb: benchmark_difficulty_level} presents the benchmark scores of each image generation model on our dataset, and \fig\ref{fig: benchmark_raderchart} shows the 18 aspect-wise scores.

As shown in \tb\ref{tb: benchmark_difficulty_level}, scores generally decrease as the action difficulty increases from \textit{Easy} to \textit{Hard}.
OMG+LoRA and OMG+InstantID achieved the highest scores at the \textit{Easy} level, with 7.28 and 7.13, respectively; however, OMG+InstantID’s score dropped notably at the \textit{Hard} level. In contrast, DreamBooth showed the smallest decrease, indicating more stable performance.

While the primary objective of our paper is to achieve human preference-aligned evaluation, the aspect-wise scores obtained as intermediate results provide detailed feedback, as illustrated in \fig\ref{fig: benchmark_raderchart}.
All models exhibited difficulty with complex actions, leading to low scores in object fidelity aspects such as \textit{Color}, \textit{Facial Similarity \& Features}, and \textit{Clothing \& Attributes}, as well as in \textit{Actions \& Expressions} and \textit{Human \& Animal Interactions}.
However, notable differences emerged in \textit{Proportions \& Body Consistency}, \textit{Object Interactions}, and \textit{Local Artifacts}, where the OMG-based methods consistently scored high across difficulty levels, while others struggled.

\subsection{Qualitative results}
\label{subsec: qualitative}

\begin{table*}[]
\centering
\caption{\textbf{Ablation studies on three components of our metric}: decomposition (Decomposition), the MLLM used for aspect-wise scoring (MLLM), and the aggregation model (Aggregation). The left value in each cell denotes Pearson’s correlation coefficient, and the right value denotes Spearman’s rank correlation.}
\label{tb: ablation}
\resizebox{0.97\textwidth}{!}{
\begin{tabular}{ccc|cccccc|c}
\hline
\textbf{Decomposition} & \textbf{MLLM} & \textbf{Aggregation} &
  \multicolumn{1}{c}{\begin{tabular}[c]{@{}c@{}}\textbf{Custom}\\\textbf{Diffusion} \end{tabular}} &
  \multicolumn{1}{c}{\begin{tabular}[c]{@{}c@{}}\textbf{OMG+}\\ \textbf{LoRA}\end{tabular}} &
  \multicolumn{1}{c}{\begin{tabular}[c]{@{}c@{}}\textbf{OMG+}\\ \textbf{InstantID}\end{tabular}} &
  \multicolumn{1}{c}{\begin{tabular}[c]{@{}c@{}}\textbf{Fast}\\ \textbf{Composer}\end{tabular}} &
  \multicolumn{1}{c}{\begin{tabular}[c]{@{}c@{}}\textbf{Mix-of-}\\ \textbf{Show}\end{tabular}} &
  \multicolumn{1}{c|}{\begin{tabular}[c]{@{}c@{}}\textbf{Dream}\\ \textbf{Booth}\end{tabular}} &
  \multicolumn{1}{c}{\textbf{Overall}} \\ \hline \hline
\xmark   & GPT-4o &  average         
& 0.77 / 0.42 & 0.42 / 0.25 & 0.50 / 0.41 & 0.51 / 0.43 & \textbf{0.67} / \textbf{0.47} & 0.58 / 0.27 & 0.67 / 0.58\\
\cmark & GPT-4o mini & average
& 0.75 / \textbf{0.56} & 0.49 / \textbf{0.42} & 0.63 / 0.44 & 0.47 / 0.28 & 0.65 / 0.37 & 0.56 / 0.29 & 0.68 / 0.59\\ 
\cmark & GPT-4o & linear regression
& 0.80 / 0.50 & 0.65 / 0.32 & 0.69 / 0.36 & 0.64 / 0.40 & 0.65 / 0.36 & \textbf{0.67} / \textbf{0.39} & 0.74 / 0.62 \\ \hline
\cmark & GPT-4o & average &   
\textbf{0.80} / 0.54 & \textbf{0.66} / 0.34 & \textbf{0.70} / \textbf{0.47} & \textbf{0.64} / \textbf{0.46} & 0.65 / 0.44 & 0.64 / 0.38 & \textbf{0.78} / \textbf{0.69} \\ \hline
\end{tabular}
}
\end{table*}

\begin{table*}[t]
\centering
\caption{\textbf{Exploration of extending our metric by combining it with existing metrics}. Ours++ denotes the combination with existing metrics. The left value in each cell represents Pearson’s correlation coefficient, and the right value represents Spearman’s rank correlation.}
\label{tb: discussion}
\resizebox{0.97\textwidth}{!}{
\begin{tabular}{l|llllll|l}
\hline
  \textbf{Model} &
  \multicolumn{1}{c}{\begin{tabular}[c]{@{}c@{}}\textbf{Custom}\\ \textbf{Diffusion}\end{tabular}} &
  \multicolumn{1}{c}{\begin{tabular}[c]{@{}c@{}}\textbf{OMG+}\\ \textbf{LoRA}\end{tabular}} &
  \multicolumn{1}{c}{\begin{tabular}[c]{@{}c@{}}\textbf{OMG+}\\ \textbf{InstantID}\end{tabular}} &
  \multicolumn{1}{c}{\begin{tabular}[c]{@{}c@{}}\textbf{Fast}\\ \textbf{Composer}\end{tabular}} &
  \multicolumn{1}{c}{\begin{tabular}[c]{@{}c@{}}\textbf{Mix-of-}\\ \textbf{Show}\end{tabular}} &
  \multicolumn{1}{c|}{\begin{tabular}[c]{@{}c@{}}\textbf{Dream}\\ \textbf{Booth}\end{tabular}} &
  \multicolumn{1}{c}{\textbf{Overall}} \\ \hline
Ours (average) 
& 0.80 / \textbf{0.54} & 0.66 / 0.34 & \textbf{0.70} / \textbf{0.47} & \textbf{0.64} / 0.46 & 0.65 / 0.44 & 0.64 / 0.38 & 0.78 / 0.69 \\
Ours++ (average) 
& 0.82 / 0.51 & \textbf{0.67} / 0.34 & 0.69 / 0.45 & 0.63 / 0.44 & 0.67 / \textbf{0.40} & \textbf{0.70} / 0.40 & \textbf{0.80} / \textbf{0.72} \\
Ours (linear regression) 
& 0.80 / 0.50 & 0.64 / \textbf{0.38} & 0.69 / 0.36 & 0.64 / 0.41 & 0.65 / 0.36 & 0.67 / 0.41 & 0.75 / 0.62 \\
Ours++ (linear regression) 
& \textbf{0.83} / 0.48 & 0.66 / 0.36 & 0.66 / 0.37 & 0.64 / \textbf{0.50} & \textbf{0.70} / 0.38 & 0.68 / \textbf{0.45} & 0.78 / 0.67 \\\hline
\end{tabular}
}
\end{table*}

\tb\ref{tb: scoring_examples} presents a scoring example on \ac{ours}.
The table shows that existing metrics produce rankings that differ considerably from human preference scores. In contrast, although minor fluctuations are observed, the proposed method generally aligns well with human preferences.

\subsection{Ablation study}
\label{subsec: ablation}
This section presents the results of ablation studies on three components of the proposed metric: the effectiveness of decomposition, the \ac{MLLM} used for aspect-wise scoring, and the aggregation model. All results are summarized in \tb\ref{tb: ablation}.

\paragraph{Decomposition ablation}
To examine the necessity of decomposition, we compared our method with a metric that directly outputs a score on a 1–10 scale using the GPT-4o model without decomposition (referred to as Vanilla-GPT).
As shown in \tb\ref{tb: ablation}, the proposed decomposed metric significantly outperforms Vanilla-GPT in both individual model scores and overall correlation with human preferences. This suggests that directly using GPT-4o without decomposition fails to capture fine-grained evaluation aspects, resulting in poor alignment with human preferences.

\paragraph{MLLM ablation}
We adopted GPT-4o as the \ac{MLLM} for aspect-wise evaluation due to its superior multi-modal understanding capabilities. However, considering API costs, we also evaluated with GPT-4o mini as a substitute.
GPT-4o mini yielded lower overall Pearson’s and Spearman’s correlations, both by 0.10, compared to GPT-4o.
Nevertheless, GPT-4o mini still outperformed existing metrics significantly (see \tb\ref{tab: correlation}). Thus, while GPT-4o provides more accurate evaluation, GPT-4o mini offers a viable alternative depending on computational resources.
Notably, the average token counts per generated image were $12,091$ (input) and $96$ (output) for GPT-4o, versus $299,672$ (input) and $104$ (output) for GPT-4o mini.

\paragraph{Aggregation model ablation}
Although this study employs averaging of aspect-wise scores to compute the final score, learnable models such as linear regression can also be used to better align with human preference scores. Results using linear regression are reported in \tb\ref{tb: ablation}.

With linear regression, the overall Pearson correlation coefficient and Spearman rank correlation coefficient are slightly lower, at 0.75 and 0.62, than those obtained with averaging.
This may be due to distribution shifts among generation models, causing slight overfitting. 
We expect that as concept customization research progresses and more generative models become available, regression-based aggregation may become more effective.

Importantly, achieving high correlation with human preferences even without regression suggests that the 18 evaluation aspects were appropriately and sufficiently defined.

%% file: latex/06_discussion.tex
In this section, we explore an extension of the proposed method by combining it with existing metrics to achieve a more comprehensive and accurate evaluation. Specifically, after obtaining 18 scores through aspect-wise evaluation, aggregation is performed by combining these scores with those from existing metrics.
\tb\ref{tb: discussion} presents the results of combining the 18 aspect-wise scores (Ours++) with scores from six existing metrics used for comparison (\ie, ArcFace, CLIP T2T score, CLIP T2I score, CLIP Aesthetic Score, DINO, and Vanilla GPT). Both the aspect-wise and existing scores were min-max normalized and scaled to a 1–10 range, resulting in a total of 24 aspects.
The table shows that combining the proposed method with existing metrics yields a slight increase in correlation compared to using the proposed method alone.

%% file: latex/07_limitation.tex
\paragraph{Dataset}
This paper's dataset excludes stationary objects and animals, focusing on humans due to the greater challenges posed by their complex structures and movements. However, since the evaluation metric also applies to scenes with objects and animals, future work will expand the dataset to include these, enabling more comprehensive evaluations.

The study focuses on generation difficulties arising from variations in human actions and emphasizes text prompt diversity. While only male and female humans were used as concept images in this study, future work will broaden the range of subjects.

\paragraph{Supported input sets}
Some image generation methods accept optional inputs, such as pose or sketch images. However, our evaluation metric assumes that generative models take only text and reference images as inputs and excludes additional inputs from the evaluation.
Considering the input cost to the \ac{MLLM}, providing all additional inputs and expanding the evaluation aspects accordingly is impractical.

Moreover, accounting for these additional inputs would increase the variability of inputs used for evaluation across models, thereby complicating accurate model comparisons.
Therefore, it is preferable to treat such additional inputs as optional during image generation, while the proposed evaluation metric focuses on the most fundamental and common inputs: text and reference images.

\paragraph{Generative models}
In this paper, we selected six generative models that support multi-concept customization as evaluation targets and obtained benchmark results. However, as discussed in \sect\ref{subsec: ablation}, due to large score variance among these models, the use of trainable aggregation methods, such as regression models, led to overfitting during training. Consequently, the correlation achieved using linear regression did not improve significantly compared to averaging. In the future, as multi-concept customization research advances and more models become available, this overfitting issue is expected to be mitigated, rendering regression a more effective aggregation method and enabling scoring with higher correlation to human preferences.

%% file: latex/08_conclusion.tex
This paper proposes \ac{our_eval}, a novel evaluation metric for concept customization.
Since existing metrics have yielded results misaligned with human preferences, we propose to decompose evaluation criteria for single and multiple concepts into 18 aspects, perform aspect-wise evaluation using \ac{MLLM}, and aggregate the results to achieve a comprehensive and human-aligned assessment.
Furthermore, we introduced \ac{ours}, a benchmark dataset supporting both single- and multi-concept evaluations.
Extensive experiments demonstrate that our metric correlates significantly better with human preferences compared to prior metrics, and ablation studies confirm the effectiveness of the decomposition strategy.

%% file: latex/99_appendix.tex
\setcounter{page}{1}
\maketitlesupplementary

\thispagestyle{empty}
\appendix
\label{sec:appendix}


\begin{table*}[h]
\caption{Comparison of existing benchmark datasets in terms of inclusion of essential components for concept customization evaluation and applicability to multi-concept customization tasks.}
\label{tab: benchmarks}
\centering
\begin{tabular}{lcccc}
\hline
 &
  \multicolumn{2}{c}{Data Type}  & 
  \multicolumn{2}{c}{Multi-Concept Support} \\ \cmidrule(lr){2-3}\cmidrule(lr){4-5} 
Dataset &
  Images &
  \multicolumn{1}{c}{text-prompt} &
  \begin{tabular}[c]{@{}c@{}}Multiple\\ Human\end{tabular} &
  \begin{tabular}[c]{@{}c@{}}Mutual\\ Interaction\end{tabular} \\ \hline\hline
MS COCO \cite{LinCOCO2014} &
  (group) & \cmark  & \cmark & \cmark \\
Flickr30K \cite{young2014From} &
  (group) & \cmark &  \cmark &  \cmark \\
FFHQ \cite{Karras2019Flickr-Faces-HQ} &
  \cmark & \cmark & \xmark & \xmark \\
UniDet \cite{zhou2021simple} &
  \cmark & \xmark & \xmark & \xmark \\
LION-400M \cite{Schuhmann2021LAION400MOD} &
  (group) & \cmark & \cmark & \xmark \\
DrawBench \cite{saharia2022photorealistic} &
  \xmark & \cmark  & \xmark & \xmark \\
$SR_{2D}$Dataset VISOR \cite{gokhale2022benchmarking} &
  \xmark & \cmark  & \cmark & \xmark \\
HRS-Bench \cite{Bakr_2023_HRS-Bench} &
  \xmark & \cmark & \cmark & \xmark \\
DreamBooth \cite{ruiz2023dreambooth} &
  \cmark & \xmark & \xmark & \xmark \\
TIFA v1.0 \cite{Hu2023TIFA} &
  \xmark & \cmark &
  \cmark &
  \xmark \\
DALLEval \cite{cho2023DALL-EVAL} & 
  (group) & \cmark & \cmark & \xmark \\
CustomConcept101 \cite{Kumari2023customdiffusion} &
  \cmark & \cmark  & \cmark & \xmark \\ 
ImagenHub \cite{ku2024imagenhub} &
  \cmark & \cmark  & \xmark & \xmark \\ 
DreamBench \cite{peng2024dreambench} &
  \cmark & \cmark  & \xmark & \xmark \\ 
  \hline
\end{tabular}
\end{table*}

\begin{table*}[t]
\centering
\caption{Action types used in \ac{ours}. \textit{Easy} and \textit{Medium} consist of the same 13 individual actions; however, in Easy, only a single person (A$^\ast$) performs the actions, whereas in Medium, two persons (A$^\ast$ and B$^\ast$) perform the same actions.
\textit{Hard} consists of 23 mutual actions.}
\label{tab: actiontypes}

\resizebox{\textwidth}{!}{
\begin{tabular}{l|l}
\hline
Individual actions (\textit{Easy / Medium})   & Mutual actions (\textit{Hard})\\ \hline \hline
\texttt{A$^\ast$ (and B$^\ast$) standing} & \texttt{A$^\ast$ punching B$^\ast$}\\
\texttt{A$^\ast$ (and B$^\ast$) walking} & \texttt{A$^\ast$ kicking B$^\ast$}\\
\texttt{A$^\ast$ (and B$^\ast$) running}  & \texttt{A$^\ast$ pushing B$^\ast$}\\
\texttt{A$^\ast$ (and B$^\ast$) waving one's hand}& \texttt{A$^\ast$ patting B$^\ast$ on the back}\\
\texttt{A$^\ast$ (and B$^\ast$) clapping} & \texttt{A$^\ast$ pointing finger at B$^\ast$}\\
\texttt{A$^\ast$ (and B$^\ast$) putting one's hands in one's pockets}  & \texttt{A$^\ast$ hugging B$^\ast$}\\
\texttt{A$^\ast$ (and B$^\ast$) jumping up}& \texttt{A$^\ast$ giving a book to B$^\ast$}\\
\texttt{A$^\ast$ (and B$^\ast$) checking the time on one's wristwatch} & \texttt{A$^\ast$ touching B$^\ast$’s pocket}\\
\texttt{A$^\ast$ (and B$^\ast$) crossing one's hands in front of one's chest}  & \texttt{A$^\ast$ and B$^\ast$ shaking hands}\\
\texttt{A$^\ast$ (and B$^\ast$) kneeling on the ground} & \texttt{A$^\ast$ hitting B$^\ast$ with a book}\\
\texttt{A$^\ast$ (and B$^\ast$) squatting down}& \texttt{A$^\ast$ putting his arm around B$^\ast$'s shoulder}\\
\texttt{A$^\ast$ (and B$^\ast$) punching} & \texttt{A$^\ast$ knocking into B$^\ast$}\\
\texttt{A$^\ast$ (and B$^\ast$) shrugging one's shoulders} & \texttt{A$^\ast$ grabbing a book from B$^\ast$}\\
& \texttt{A$^\ast$ stepping on B$^\ast$'s foot}\\
& \texttt{A$^\ast$ and B$^\ast$ giving each other a high-five}\\
& \texttt{A$^\ast$ and B$^\ast$ clinking glasses}\\
& \texttt{A$^\ast$ and B$^\ast$ carrying a box together}\\
& \texttt{A$^\ast$ taking a picture of B$^\ast$ with a camera}\\
& \texttt{A$^\ast$ following B$^\ast$ down a street}\\
& \texttt{A$^\ast$ whispering into B$^\ast$’s ear}\\
& \texttt{A$^\ast$ and B$^\ast$ exchanging a book}\\
& \texttt{A$^\ast$ supporting B$^\ast$ as they walk}\\
& \texttt{A$^\ast$ and B$^\ast$ playing rock-paper-scissors}\\\hline
\end{tabular}
}
\end{table*}

\begin{table*}[]
\centering
\caption{Prompt templates used for image generation in our dataset}
\label{tab: image_generation_prompt}
\small
\begin{tabular}{ll}
\hline
\multicolumn{2}{l}{Prompting template for the base image generation} \\ \hline
A Woman &
  \begin{tabular}{p{14cm}}
  \texttt{Generate a high-quality, photo-realistic, full-body image (including the legs) of a woman as described below:}\\ 
  \texttt{- Hairstyle: long blonde hair}\\ 
  \texttt{- skin tone: white}\\ 
  \texttt{- Age: around 25 years old}\\ 
  \texttt{- Face: smiling at the camera}\\ 
  \texttt{- Outfit: casual wearing necklace}\\ 
  \texttt{- Output a square image}\end{tabular}
   \\ \hline
A Man &
  \begin{tabular}{p{14cm}}
  \texttt{Generate a high-quality, photo-realistic, full-body image (including the legs) of a man as described below:}  \\   
  \texttt{- Hairstyle: black hair}\\   
  \texttt{- skin tone: brown}  \\   
  \texttt{- Age: around 25 years old}  \\   
  \texttt{- Face: smiling at the camera}  \\   
  \texttt{- Outfit: in his suite, with a bag}\\   
  \texttt{- Output a square image}\end{tabular} \\ \hline
\multicolumn{2}{l}{Prompting example for image replication}          \\ \hline
\multicolumn{2}{l}{\begin{tabular}[c]{@{}l@{}}
\texttt{Generate close-up photo of this man / woman who looks lonely}\\ 
\texttt{Generate side-shot photo of this man / woman in the image}\\ 
\texttt{Generate close-up photo of him / her, who is thinking deeply with serious face}\end{tabular}} \\ \hline
\end{tabular}
\end{table*}

\newpage
\begin{table*}[]
\vspace{1cm}
\centering
\caption{\textbf{Prompting template for different types of inputs to GPT}. For each evaluation aspect, we selectively determined whether to provide the text prompt $\mathcal{T}$, the reference images $\mathcal{I}$, or neither with the generated image $I_g$, and accordingly made slight adjustments to the prompting templates.}
\label{tab: prompt_to_gpt_ours}
\small

\begin{tabular}{c|c|c|l}
\hline
\textbf{$I_g$}&
  \textbf{$\mathcal{T}$}&
  \textbf{$\mathcal{I}$} &
  \multicolumn{1}{c}{\textbf{Prompting Template}} \\ \hline
\cmark &
  \cmark &
   &
  \begin{tabular}{p{13cm}}\texttt{Task:}\\ 
  \texttt{I will provide a text prompt, followed by a generated image. Please rate how well the generated image meets the following evaluation aspect, then give a score from 1 to 5. DO NOT check whether the generated image matches the entire text prompt. Instead, rate it solely based on the following evaluation aspect.}\\ \\ 
  \texttt{Evaluation aspect:}\\ 
  \textless{}\textit{Evaluation aspect}\textgreater\\ \\ 
  \texttt{Scoring example:}\\ 
  \textless{}\textit{Scoring example}\textgreater\\ \\
  \texttt{The text prompt:}\\ 
  ``\textless{}\textit{Text prompt}\textgreater{}''\\ \\ \textless{}\textit{Generated image}\textgreater\\ \\ \texttt{Score:}
  \end{tabular} \\\hline
\cmark &
  &
  \cmark &
  \begin{tabular}{p{13cm}}\texttt{Task:}\\
  \texttt{I will provide a generated image, followed by one or two reference images. Please observe carefully how well the generated image meets the following evaluation aspect, then give a score from 1 to 5.}\\ \\ \\ 
  \texttt{Evaluation aspect:}\\ 
  \textless{}\textit{Evaluation aspect}\textgreater\\ \\
  \texttt{Scoring example:}\\ 
  \textless{}\textit{Scoring example}\textgreater\\ \\ \textless{}\textit{Generated image, reference images}\textgreater\\ \\ 
  \texttt{Score:}\end{tabular} \\ \hline
\cmark &
   &
   &
  \begin{tabular}{p{13cm}}\texttt{Task:}\\
  \texttt{I will present a set of images cropped at different locations of the same generated image. Please observe carefully how well the generated image meets the following evaluation aspect, then give a score from 1 to 5 based on the worst image.}\\ \\ 
  \texttt{Evaluation aspect:}\\ 
  \textless{}\textit{Evaluation aspect}\textgreater\\ \\ 
  \texttt{Scoring example:}\\ 
  \textless{}\textit{Scoring example}\textgreater\\ \\
  \textless{}\textit{generated image}\textgreater\\ \\ \texttt{Score:}\end{tabular} \\ \hline
\end{tabular}
\vspace{1cm}
\end{table*}

\section{Existing Benchmark for Concept Customization}
\tb\ref{tab: benchmarks} summarizes the comparison of the composition of existing benchmark datasets used for concept customization evaluation.

\section{Dataset Details}

\subsection{Dataset License}
\label{apx_sec: dataset_license}
We released our benchmark dataset, \ac{ours}, and related materials under the CC BY-NC 4.0.

\subsection{Text generation details}
\label{apx_sec: dataset_action_types}

In \tb\ref{tab: actiontypes}, we list all the action types used in our dataset.
The action types are consistent between the \textit{Easy} and \textit{Medium} levels. A portion of the action types used in the dataset are adapted from \cite{Shahroudy2016NTURGB, Liu2020NTURGB+D120}.

The surroundings specified in the prompts are drawn from 20 distinct types, which are reused across different prompts.

\subsection{Image generation details}
\label{apx_sec: im_gen_prompt}


\begin{figure}[]
    \centering
        \includegraphics[width=\linewidth]{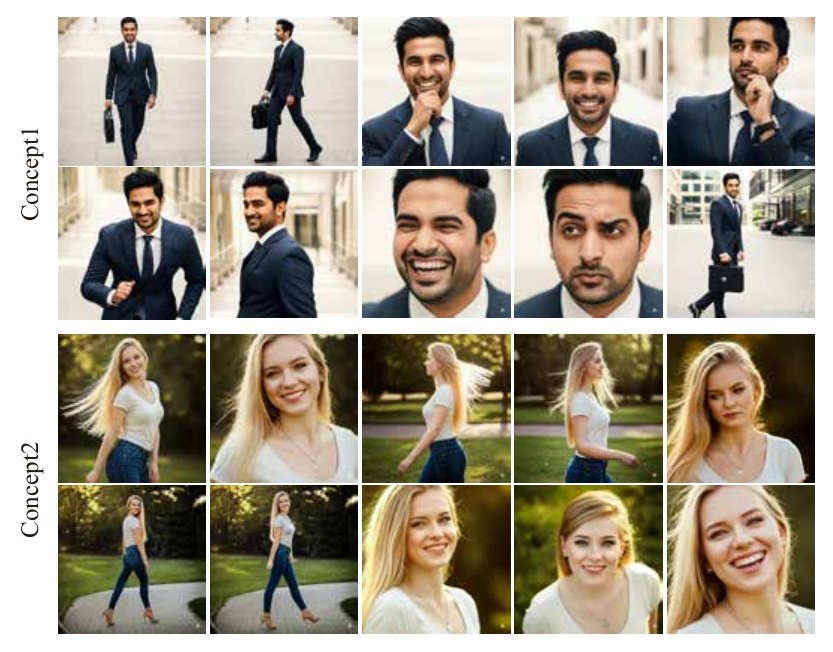}
        \caption{\textbf{Sample images in \ac{ours}}}
        \label{fig: sample_dataset}
\end{figure}

We generated images using the text prompts listed in \tb\ref{tab: image_generation_prompt}.
Regarding the alignment between the text prompts and the resulting images, strict adherence was not prioritized. Instead, emphasis was placed on introducing variation, with visual inspection confirming that the subject's appearance remained consistent with the original image.
Figure \ref{fig: sample_dataset} presents ten samples generated for each concept.

\section{Approach to Defining Evaluation Aspects}
\label{apx_sec: identification_of_eval_aspects_details}

In determining the 18 evaluation aspects, we adopted a bottom-up categorization approach. This study aims to achieve human-aligned evaluation based on the premise that humans can comprehensively perceive all relevant factors. Accordingly, we first exhaustively identified potential perspectives for evaluating concept customization images. Considering the cost of using \ac{MLLM}, we then minimized the number of aspects while maintaining appropriate evaluation granularity by removing redundant items and grouping related ones.
The overall categorization was inspired by the evaluation aspects proposed in prior studies~\cite{zhong2024multilora, He2024VideoScore}.

\section{Prompting of Aspect-Wise Evaluation}
\label{apx_sec: detail_of_aspect-wise_eval}

To obtain the aspect-wise scores, we provide the \ac{MLLM} model (\ie, the GPT model) with the generated image $I_g$, the text prompt $\mathcal{T}$, and the reference images $\mathcal{I}$ as inputs.
\tb\ref{tab: prompt_to_gpt_ours} presents the prompting template for the MLLM model, while the prompts that vary depending on the evaluation aspect are detailed in \tb\ref{tab: aspect-wise_eval_details}.

It has been empirically demonstrated that by eliminating unnecessary inputs, \ac{MLLM} can provide more accurate evaluations and reduce the associated costs. Therefore, we evaluated the necessity of the text prompt and reference images for each evaluation aspect. For instance, when evaluating aspects that only compare the consistency between the generated image and the text prompt, reference images are not provided to the \ac{MLLM}.
\tb\ref{tab: aspect-wise_eval_details} indicates which input---text prompt, reference images, or neither---is used for each evaluation aspect, and \tb\ref{tab: prompt_to_gpt_ours} highlights the template differences between various input combinations.

\section{Prompting of Vanilla-GPT}
As an ablation study to evaluate the effect of decomposition, we conducted an experiment in which images were rated on a scale from 1 to 10 by directly prompting GPT-4o without applying decomposition.
The prompting template used for GPT-4o in this experiment is provided in \tb\ref{tab: prompting_template_of_vanillagpt}.

\section{Experimental Details}
\label{apx_sec: experimental_details}

\subsection{Details of existing metrics}
\label{apx_subsec: existing_evaluation_metrics}

\paragraph{ArcFace} utilizes face detection methods such as MTCNN \cite{Zhang2016mtcnn} to localize faces and measures feature similarity within the embedding space \cite{Xiao2024FastComposer, he2024uniportrait, Kong2024OMG}.
In this study, faces detected with high confidence by MTCNN are used to extract embedding features for the specified number of concepts via Inception ResNet (V1) pretrained on VGGFace2~\cite{szegedy2014goingdeeperconvolutions, Cao2018VGGFace2}.
Embedding features are similarly extracted from reference images, and their similarities are computed.
For multi-concept cases, the highest similarity per embedding is selected, and the average across all embeddings forms the final score.
If fewer faces are detected than concepts specified, missing face scores are set to zero.

\paragraph{CLIP T2T} automatically generates captions from generated images using BLIP-2 \cite{li2023blip2}, and then calculates the similarity of CLIP embedding features between the input text and the generated caption \cite{Kumari2023customdiffusion, Liu2023Cones2, Li2023BLIPdiffusion, Ma2024Subject-Diffusion}. In our investigation, this metric is currently the most popular evaluation method for concept customization.

\paragraph{CLIP T2I} calculates the similarity of CLIP embedding features between the input text and the generated image \cite{Kumari2023customdiffusion, Kong2024OMG, Ma2024Subject-Diffusion}.
This method is one of the most commonly used evaluation approaches, following CLIP T2T.

\paragraph{DINO Score} is a method that inputs the generated image along with the text prompt or reference image into DINO's Vision Transformer encoder \cite{Caron2021Dino}, and calculates the similarity of the resulting feature vectors \cite{ruiz2023dreambooth, Chen2024AnyDoor, Ma2024Subject-Diffusion, Li2023BLIPdiffusion}.

\paragraph{CLIP Aesthetic Score} uses a CLIP model fine-tuned on an aesthetic evaluation dataset to calculate aesthetic scores from the embedding features \cite{schuhmann2022laion5bopenlargescaledataset}.
In this study, we adopted the \texttt{LAION-Aesthetics Predictor V1}\footnote{\url{https://github.com/LAION-AI/aesthetic-predictor}}.

\subsection{Implementation details of generative models}
\label{apx_subsec: generative_models}

\paragraph{CustomDiffusion}
For each concept tuning, 20 images were used, and the model was trained for 3000 steps with a batch size of 2 and a learning rate of 5e-6, after which it was adapted for multi-concept use. During both training and sampling, the cross-attention parameters were frozen.

\paragraph{OMG+LoRA}
To train the character LoRA, we utilized the Kohya\_ss \footnote{\url{https://github.com/bmaltais/kohya_ss}} with a learning rate of 0.003, the Adafactor optimizer, a rank of 256, a batch size of 4, and 5 epochs, following the experimental setup of the original paper.
For visual comprehension, we employed the GroundingDINO~\cite{Liu2024GroundingDINO} combined with SAM~\cite{Kirillov2023sam}.
The negative prompt used was: ``noisy, blurry, soft, deformed, ugly.''

\paragraph{OMG+InstantID}
Similar to OMG+LoRA, we employed GroundingDINO~\cite{Liu2024GroundingDINO} combined with SAM~\cite{Kirillov2023sam} for visual comprehension, using the same negative prompt: ``noisy, blurry, soft, deformed, ugly.''
Since this method requires only a single image per concept for sampling, one full-body image was selected for each concept.

\paragraph{FastComposer}
For each concept, a single full-body image was selected. The parameters guidance\_scale, inference\_steps, and start\_merge\_step were set to their default values of 5, 50, and 10, respectively. No additional retraining or fine-tuning was performed.

\paragraph{Mix-of-Show}
For tuning the embedding-decomposed LoRA (ED-LoRA) per concept, all 20 images for each concept were used, and SAM was employed to generate the required mask images for training.
In multi-concept settings, both unet\_alpha and text\_encoder\_alpha were set to 1.0 as the default for concept fusion.
Although keypose images or sketches can be used in the multi-concept sampling process, these inputs were disabled to ensure consistency with other methods. Since layout was the only mandatory aspect, the input consisted solely of a split layout, dividing the screen into left and right sections.
The negative prompt used was the following, as specified in the original method’s proposal: ``longbody, lowres, bad anatomy, bad hands, missing fingers, extra digit, fewer digits, cropped, worst quality, low quality.''

\paragraph{DreamBooth}
The implementation of the proposed model adopts Diffuser’s Multi-Subject DreamBooth~\cite{platen2022diffusers}. All parameters were kept at their default settings, and for each concept, training was conducted using all 20 images in our dataset, with a batch size of 1, 1500 training steps, and a learning rate of 1e-6.

\section{Annotation Details}
\label{apx_sec: user study}

\begin{figure*}[]
  \includegraphics[width=\textwidth]{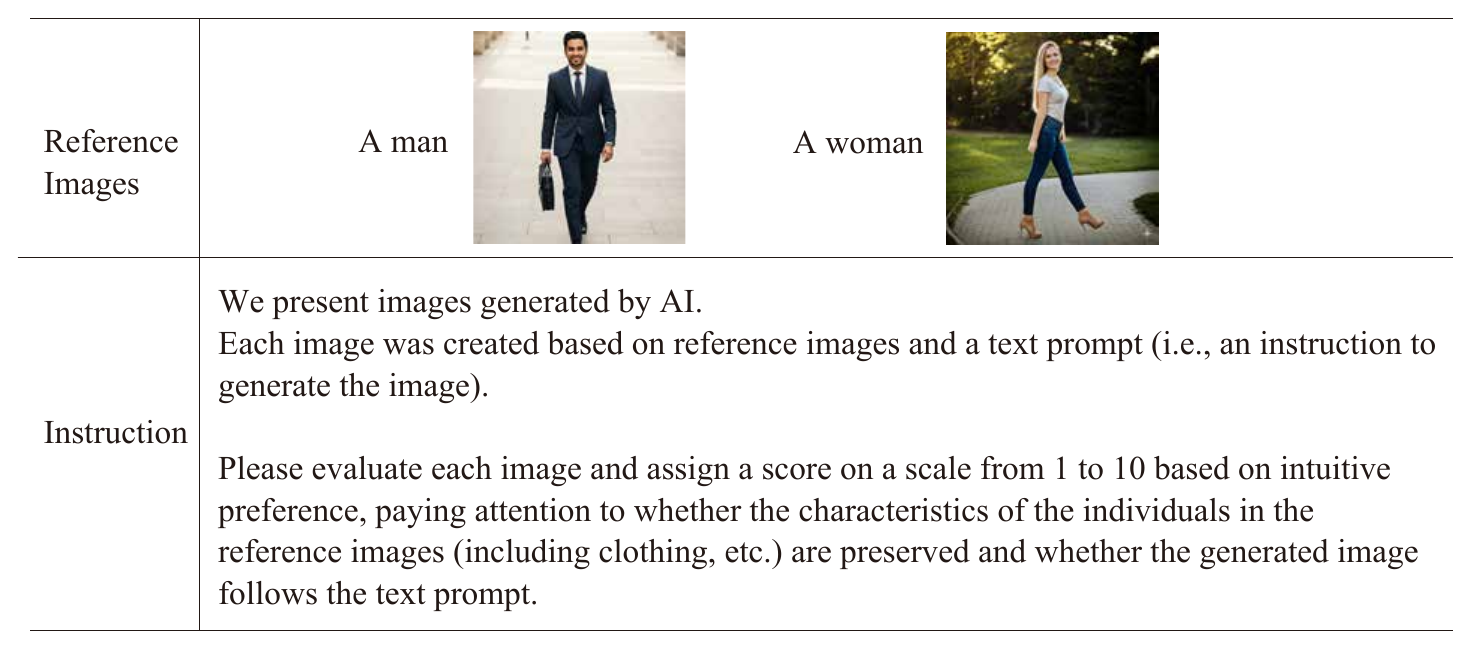}
\caption{Instruction of annotation}
\label{fig: instruction}
\end{figure*}

Detailed instructions for the annotation process are provided in \fig\ref{fig: instruction}. All annotators were informed about the purpose of this dataset and provided consent before participation.
The purpose of this annotation was to measure the correlation between human intuitive evaluations and the results obtained by the proposed method. 
To avoid bias introduced by detailed evaluation criteria, only the minimum necessary information for evaluating concept customization was provided. 
Detailed evaluation aspects were intentionally omitted to encourage annotators to rely on their own judgment criteria.

\section{Scatter Plot of Predicted Scores}
\label{apx_sec: scatter_plot}

\begin{figure*}[]
  \includegraphics[width=\textwidth]{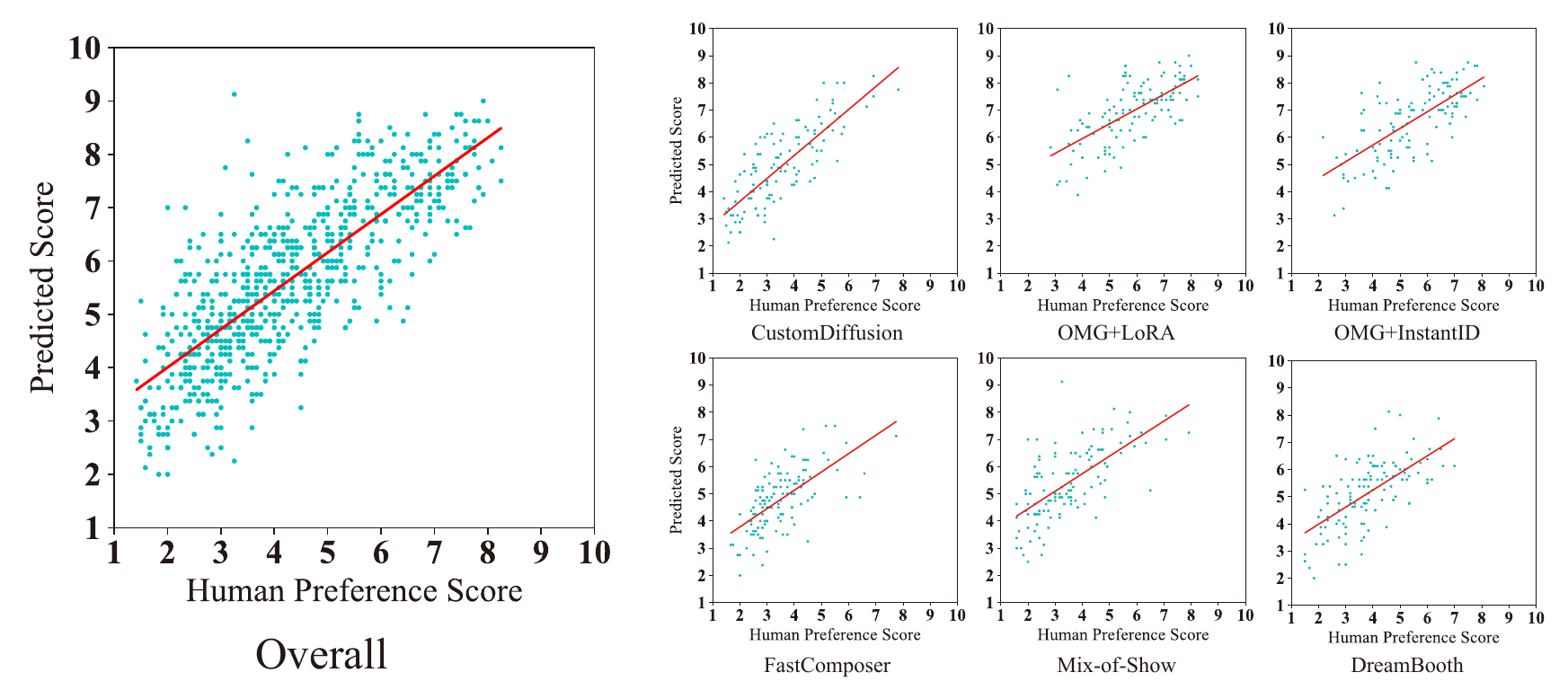}
  \caption{Scatter plots of the human preference scores versus our predicted scores.}
  \label{fig: plot}
\end{figure*}

\fig\ref{fig: plot} presents scatter plots of the human preference scores versus our predicted scores, shown both for all models collectively and for each model individually.
The red line represents the regression line.
From the figure, it can be observed that, for each method, the human preference scores and predicted scores exhibit a strong correlation.

\section{Ranking Comparison}
\tb\ref{tab: ranking} presents the average ranking of each metric's scores across models.
Existing methods exhibit considerable variation in their deviation from human preference based on average rankings. In contrast, our proposed method consistently achieves a ranking difference of less than 1 from human preference across all generative models, demonstrating stable alignment with human preference.

\section{Case Study}
\label{apx_sec: case_study}

\begin{figure*}[]
\centering
  \includegraphics[width=\textwidth]{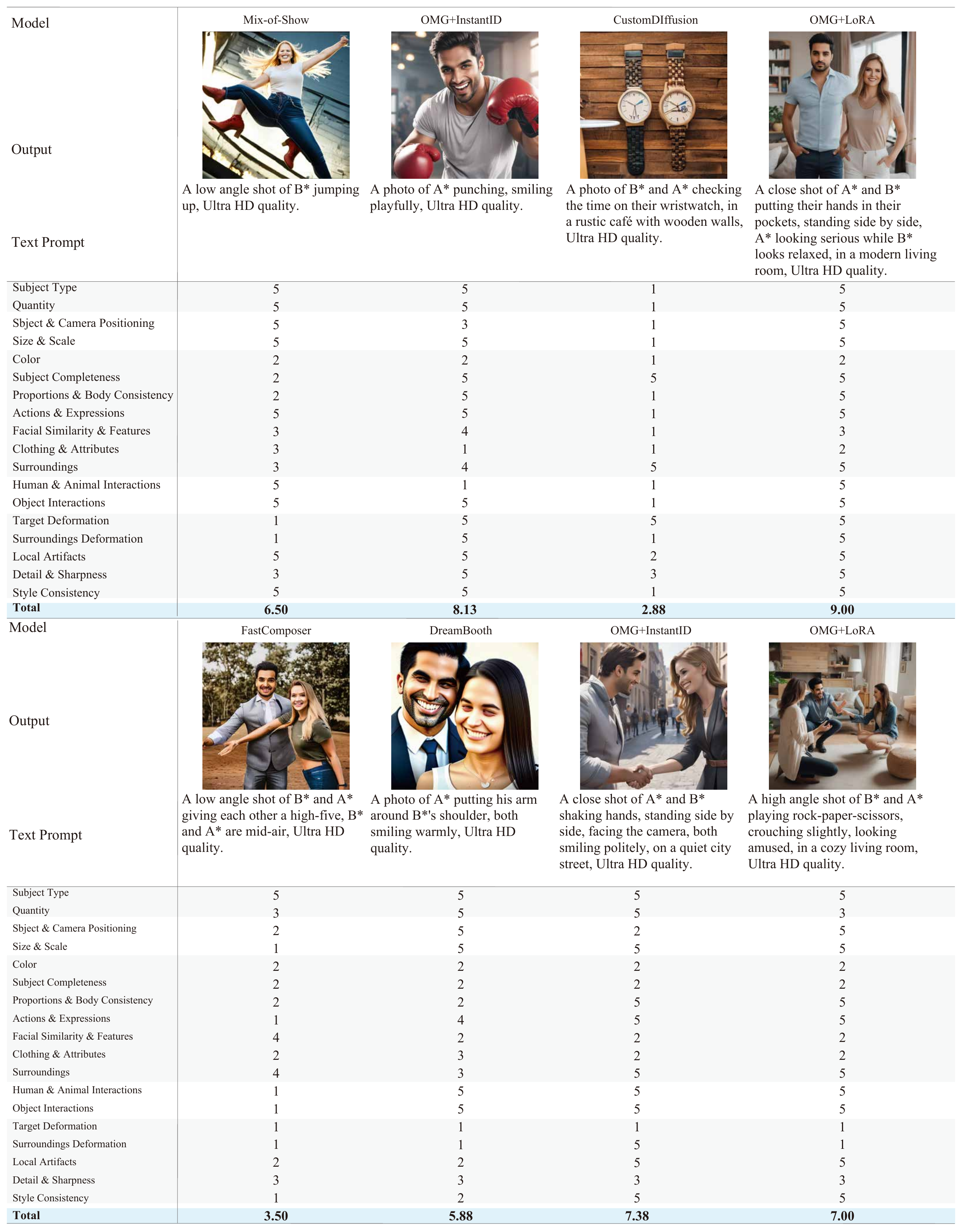}
\caption{Case study}
\label{fig: case_study}
\end{figure*}

\fig\ref{fig: case_study} shows some examples of our aspect-wise scores and their aggregated scores.

\clearpage


\begin{table*}[]
\centering
\caption{\textbf{Details of the input to GPT for each evaluation aspect}. Each evaluation aspect is fed into the model along with the generated image $I_g$, and either the text prompt $\mathcal{T}$ and reference images $\mathcal{I}$, or neither.}
\label{tab: aspect-wise_eval_details}

\small
\begin{tabular}{l|c|c|c|l}
\hline
\begin{tabular}{p{1.5cm}}Evaluation Aspect\end{tabular} &
  $I_g$&
  $\mathcal{T}$&
  $\mathcal{I}$ &
  \begin{tabular}{p{10cm}}Prompt\end{tabular}\\\hline
  
\begin{tabular}{p{1.5cm}}Subject \\Type\end{tabular} &
  \cmark &
  \cmark &
   &
  \begin{tabular}{p{11cm}}\texttt{Evaluation aspect:}\\ 
  \texttt{Do the generated objects and people match the specified types (\eg, 'a man' should not be misrepresented as 'a woman')?  Focus ONLY on the subject type accuracy.}\\ \\ 
  \texttt{Scoring example:}\\ 
  \texttt{If the genders are swapped, subtract 4 points.}\end{tabular} \\ \hline
  
Quantity &
  \cmark &
  \cmark &
   &
  \begin{tabular}{p{11cm}}\texttt{Evaluation aspect:}\\ \texttt{Are the correct number of objects and persons generated according to the prompt? Focus ONLY on quantity accuracy. }\\ \\ 
  \texttt{Scoring example:}\\ 
  \texttt{If the prompt specifies two men but three are generated, subtract 4 points.}\end{tabular} \\ \hline
  
\begin{tabular}{p{1.5cm}}
Subject \& \\Camera \\Positioning\end{tabular} &
  \cmark &
  \cmark &
   &
  \begin{tabular}{p{11cm}}\texttt{Evaluation aspect:}\\ \texttt{Are objects and people positioned correctly and arranged logically within the scene, preserving appropriate spatial relationships, depth, and occlusion according to the specified layout? Focus ONLY on the subject and camera positioning. If there is no relevant part in the text prompt, ignore the prompt.}\\ \\ \texttt{Scoring example:}\\ \texttt{If a 'long shot' is required but a close-up is generated, subtract 3 points.}\end{tabular} \\ \hline
  
\begin{tabular}{p{1.5cm}}Size \& \\Scale \end{tabular} &
  \cmark &
  \cmark &
   &
  \begin{tabular}{p{11cm}}\texttt{Evaluation aspect:}\\ \texttt{Are the absolute and relative sizes of objects and people appropriate for the scene? Focus ONLY on the size and scale. }\\ \\ 
  \texttt{Scoring example:}\\ 
  \texttt{If the man in the image appears too small relative to the surrounding objects, subtract 4 points.}\end{tabular} \\ \hline
  
Color &
  \cmark &
   &
  \cmark &
  \begin{tabular}{p{11cm}}\texttt{Evaluation aspect:}\\ \texttt{Are the colors applied appropriately according to the reference images? Focus ONLY on the color accuracy.}\\ \\ \texttt{Scoring example:}\\ \texttt{If the skin tone or hair color is different from the reference image, subtract 3 points.}\end{tabular} \\ \hline
  
\begin{tabular}{p{1.5cm}} Subject Completeness\end{tabular} &
  \cmark &
   &
   &
  \begin{tabular}{p{11cm}}\texttt{Evaluation aspect:}\\ \texttt{Is the object or person fully generated with no missing or extra parts? Focus ONLY on the subject completeness.}\\ \texttt{*Pay special attention to where the two individuals are in contact*}\\ \\ 
  \texttt{Scoring example:}\\ 
  \texttt{If the hands touching the other person are semi-transparent or unclear, subtract 3 points.}\end{tabular} \\ \hline
  
\end{tabular}
\end{table*}

\clearpage

\begin{table*}[t]
\centering
\small
\begin{tabular}{l|c|c|c|l}
\hline

\begin{tabular}{p{1.5cm}}Evaluation Aspect\end{tabular} &
  $I_g$&
  $\mathcal{T}$&
  $\mathcal{I}$ &
  \begin{tabular}{p{10cm}}Prompt\end{tabular}\\\hline

    \begin{tabular}{p{1.5cm}} Proportions \\\& Body \\Consistency\end{tabular} &
  \cmark &
   &
  \cmark &
  \begin{tabular}{p{11cm}}\texttt{Evaluation aspect:}\\ \texttt{Are body proportions and limb positioning natural and consistent with the given text prompt or reference image? Focus ONLY on the proportions and body consistency.}\\ \\ 
  \texttt{Scoring example:}\\ 
  \texttt{If the limb or arm is unnatural, subtract 4 points; if the body proportions are off, subtract 3 points.}\end{tabular} \\ \hline
  
   \begin{tabular}{p{1.5cm}} Actions \& \\Expressions\end{tabular} &
  \cmark &
  \cmark &
   &
  \begin{tabular}{p{11cm}}\texttt{Evaluation aspect:}\\ \texttt{Are specified actions, poses, gaze direction, and facial expressions correctly depicted, reflecting the intended motion and emotion from the text prompt? Focus ONLY on the actions and expressions.}\\ \\ 
  \texttt{Scoring example:}\\ 
  \texttt{If the man is instructed to laugh but isn't, subtract 4 points.}\end{tabular} \\ \hline
\begin{tabular}{p{1.5cm}} Clothing \& Attributes\end{tabular} &
  \cmark &
   &
  \cmark &
  \begin{tabular}{p{11cm}}\texttt{Evaluation aspect:}\\ \texttt{Are clothing, accessories, and key features consistent with the reference images? Focus ONLY on the clothing and attributes.}\\ \\ 
  \texttt{Scoring example:}\\ 
  \texttt{If the person is missing accessories, subtract 1 point; if the clothing differs completely from the reference, subtract 2 points.}\end{tabular} \\ \hline

  \begin{tabular}{p{1.5cm}} Facial Similarity \& Features\end{tabular} &
  \cmark &
   &
  \cmark &
  \begin{tabular}{p{11cm}}\texttt{Evaluation aspect:}\\ \texttt{Does the generated face resemble the reference image, preserving key characteristics like shape, expression, and symmetry? Focus ONLY on the facial similarity and features.}\\ \\ 
  \texttt{Scoring example:}\\ 
  \texttt{If the face differs from the reference but keeps key features like hairstyle, subtract 3 points.}\end{tabular} \\ \hline
  
\begin{tabular}{p{1.5cm}} Surroundings \end{tabular} &
  \cmark &
  \cmark &
   &
  \begin{tabular}{p{11cm}}\texttt{Evaluation aspect:}\\ 
  \texttt{Is the surrounding environment accurately depicted according to the provided text prompt? Focus ONLY on the surroundings.}\\ \\ 
  \texttt{Scoring example:}\\ 
  \texttt{If a cafe is specified but a photo of a park is generated, assign 1 point; if there is no relevant part in the text prompt, ignore the prompt.}\end{tabular} \\ \hline
  
\begin{tabular}{p{1.5cm}} Human \&\\Animal Interactions\end{tabular} &
  \cmark &
  \cmark &
   &
  \begin{tabular}{p{11cm}}\texttt{Evaluation aspect:}\\ \texttt{Are persons and animals interacting naturally with objects and each other as specified in the text prompt? Focus ONLY on the human and animal interactions.}\\ \\ 
  \texttt{Scoring example:}\\ 
  \texttt{If the prompt specifies hugging but the image shows handshaking, subtract 4 points.}\end{tabular} \\ \hline

\end{tabular}
\end{table*}


\begin{table*}[t]
\centering
\small
\begin{tabular}{l|c|c|c|l}
\hline
\begin{tabular}{p{1.5cm}}Evaluation Aspect\end{tabular} &
  $I_g$&
  $\mathcal{T}$&
  $\mathcal{I}$ &
  \begin{tabular}{p{10cm}}Prompt\end{tabular}\\\hline

  \begin{tabular}{p{1.5cm}} Object Interactions\end{tabular} &
  \cmark &
  \cmark &
   &
  \begin{tabular}{p{11cm}}\texttt{Evaluation aspect:}\\ \texttt{Are objects interacting logically within the scene as specified in the text prompt? Focus ONLY on the object interactions.}\\ \\ 
  \texttt{Scoring example:}\\ 
  \texttt{If the book in the prompt sinks into the table, subtract 4 points.}\end{tabular} \\ \hline

  \begin{tabular}{p{1.5cm}} Subject \\Deformation\end{tabular} &
  \cmark &
   &
   &
  \begin{tabular}{p{11cm}}\texttt{Evaluation aspect:}\\ \texttt{Are the people in the image (especially the faces and where the two individuals are in contact) rendered without deformations? Focus ONLY on the subject's deformation. }\\ \\ 
  \texttt{Scoring example:}\\ 
  \texttt{If the person's face has any deformation or unrecognizable, subtract 4 points.}\end{tabular} \\ \hline
  
  \begin{tabular}{p{1.5cm}} Surroundings Deformation\end{tabular} &
  \cmark &
   &
   &
  \begin{tabular}{p{11cm}}\texttt{Evaluation aspect:}\\ \texttt{Are the surroundings in the image rendered naturally, without deformations such as crooked lines or unnatural parts? Focus ONLY on the surroundings' deformation.}\\ \\ 
  \texttt{Scoring example:}\\ 
  \texttt{If the surroundings have deformation, subtract 4 points.}\end{tabular} \\ \hline
  
    \begin{tabular}{p{1.5cm}} Local \\Artifacts\end{tabular} &
  \cmark &
   &
   &
  \begin{tabular}{p{11cm}}\texttt{Evaluation aspect:}\\ \texttt{Are the image rendered without unwanted noise, strange patterns, or incomplete renderings? Focus ONLY on the local artifacts.}\\ \\ 
  \texttt{Scoring example:}\\ 
  \texttt{If there is an unwanted watermark on the generated image, subtract 3 points.}\end{tabular} \\ \hline

  \begin{tabular}{p{1.5cm}} Detail \& \\Sharpness\end{tabular} &
  \cmark &
   &
   &
  \begin{tabular}{p{11cm}}\texttt{Evaluation aspect:}\\ \texttt{Are facial features, hands, and intricate details well-defined? Focus ONLY on the detail and sharpness.}\\ \\ 
  \texttt{Scoring example:}\\ 
  \texttt{If the entire image lacks detail, subtract 4 points; if a person is missing detail in any part (\eg, hands, legs, arms, face), subtract 2 points.}\end{tabular} \\ \hline

  \begin{tabular}{p{1.5cm}} Style \\Consistency\end{tabular} &
  \cmark &
  \cmark &
   &
  \begin{tabular}{p{11cm}}\texttt{Evaluation aspect:}\\ \texttt{Does the generated image adhere to the artistic or visual style specified in the text prompt? Focus ONLY on the style consistency.}\\ \\ 
  \texttt{Scoring example:}\\ 
  \texttt{If the prompt requires a realistic image but the style is anime-like, subtract 4 points.}\end{tabular} \\ \hline

\end{tabular}
\end{table*}

\begin{table*}[htbp]
\centering
\caption{Prompting template of Vanilla-GPT}
\label{tab: prompting_template_of_vanillagpt}
\begin{tabular}{c|c|c|l}
\hline
\textbf{$I_g$}&
  \textbf{$\mathcal{T}$}&
  \textbf{$\mathcal{I}$} &
  \multicolumn{1}{c}{\textbf{Prompting Template}} \\ \hline
\cmark &
  \cmark &
  \cmark &
  \begin{tabular}{p{13cm}}\texttt{Task:}\\
  \texttt{I will provide a text prompt, followed by a generated image and one or two reference images. Please evaluate the generated image 
  and assign a score on a scale from 1 to 10. Pay attention to whether the characteristics of the individuals in the reference 
  images (including clothing, etc.) are preserved and whether the generated image follows the text prompt.}\\ \\ \\ 
  \texttt{The text prompt}\\ 
  ``\textless{}\textit{Text prompt}\textgreater{}'' \\ \\ \textless{}\textit{Generated image, reference images}\textgreater\\ \\ \texttt{Score:}\end{tabular} \\ \hline
\end{tabular}
\end{table*}

\clearpage


\begin{table*}[]
\centering
\caption{\textbf{Average rank of scores across models}.
The metric closest to human preference is highlighted in \textbf{bold}, and the second closest metric is indicated by an \underline{underline}.}
\label{tab: ranking}
\resizebox{\linewidth}{!}{
\begin{tabular}{l|cccccc}
\hline
Metric 
& \multicolumn{1}{c}{\begin{tabular}[c]{@{}c@{}}CustomDiffusion\end{tabular}}
& \multicolumn{1}{c}{\begin{tabular}[c]{@{}c@{}}OMG\\+LoRA\end{tabular}} 
& \multicolumn{1}{c}{\begin{tabular}[c]{@{}c@{}}OMG\\+InstantID\end{tabular}} 
& \multicolumn{1}{c}{\begin{tabular}[c]{@{}c@{}}FastComposer\end{tabular}} 
& \multicolumn{1}{c}{\begin{tabular}[c]
{@{}c@{}}Mix-of-Show\end{tabular}}  
& \multicolumn{1}{c}{\begin{tabular}[c]{@{}c@{}}DreamBooth\end{tabular}} 
\\ \hline \hline
ArcFace   
& 4.50 \scriptsize{(-0.14)}
& 3.41 \scriptsize{(-1.83)}
& 3.75 \scriptsize{(-1.73)}
& 1.88 \scriptsize{(+2.55)}
& 2.64 \scriptsize{(+1.88)}
& 4.56 \scriptsize{(-0.68)}\\
CLIP T2I  
& 3.15 \scriptsize{(+1.21)}
& 2.62 \scriptsize{(-1.04)}
& 2.61 \scriptsize{(-0.59)}
& \underline{4.58} \scriptsize{(-0.15)}
& \textbf{4.53} \scriptsize{(-0.01)}
& \underline{3.51} \scriptsize{(+0.37)}\\
CLIP T2T  
& 3.49 \scriptsize{(+0.87)}
& 3.52 \scriptsize{(-1.94)}
& 2.63 \scriptsize{(-0.61)}
& 3.89 \scriptsize{(+0.54)}
& 4.05 \scriptsize{(+0.47)}
& 3.37 \scriptsize{(+0.51)}\\
CLIP Aes. 
& 5.14 \scriptsize{(-0.78)}
& 2.39 \scriptsize{(-0.81)}
& 1.72 \scriptsize{(+0.30)}
& 3.86 \scriptsize{(+0.57)}
& \underline{4.60} \scriptsize{(-0.08)}
& 3.29 \scriptsize{(+0.59)}\\
DINO      
& \textbf{4.31} \scriptsize{(+0.05)}
& 3.40 \scriptsize{(-1.82)}
& 3.49 \scriptsize{(-1.47)}
& 3.27 \scriptsize{(+1.16)} 
& 3.11 \scriptsize{(+1.41)}
& 3.42 \scriptsize{(+0.46)}\\ 
Vanilla-GPT 
& 4.23 \scriptsize{(+0.13)}
& \textbf{1.66} \scriptsize{(-0.08)}
& \textbf{1.99} \scriptsize{(+0.03)}
& 3.29 \scriptsize{(+1.14)}
& 3.58 \scriptsize{(+0.94)}
& 3.14 \scriptsize{(+0.74)}\\ \hline
Ours      
& \underline{4.30} \scriptsize{(+0.06)}
& \underline{1.68} \scriptsize{(-0.10)}
& \underline{2.16} \scriptsize{(-0.14)}
& \textbf{4.56} \scriptsize{(-0.13)}
& 3.79 \scriptsize{(+0.73)}
& \textbf{4.13} \scriptsize{(-0.25)}\\ \hline
Human Preference & \textbf{4.36} & \textbf{1.58} & \textbf{2.02} & \textbf{4.43} & \textbf{4.52} & \textbf{3.88} \\\hline

\end{tabular}
}
\end{table*}


%% file: main.bbl
\begin{thebibliography}{63}
\providecommand{\natexlab}[1]{#1}
\providecommand{\url}[1]{\texttt{#1}}
\expandafter\ifx\csname urlstyle\endcsname\relax
  \providecommand{\doi}[1]{doi: #1}\else
  \providecommand{\doi}{doi: \begingroup \urlstyle{rm}\Url}\fi

\bibitem[Bakr et~al.(2023)Bakr, Sun, Shen, Khan, Li, and Elhoseiny]{Bakr_2023_HRS-Bench}
Eslam~Mohamed Bakr, Pengzhan Sun, Xiaoqian Shen, Faizan~Farooq Khan, Li~Erran Li, and Mohamed Elhoseiny.
\newblock {HRS-Bench: Holistic, Reliable and Scalable Benchmark for Text-to-Image Models}.
\newblock In \emph{ICCV}, 2023.

\bibitem[Bansal et~al.(2025{\natexlab{a}})Bansal, Lin, Xie, Zong, Yarom, Bitton, Jiang, Sun, Chang, and Grover]{bansal2024videophy}
Hritik Bansal, Zongyu Lin, Tianyi Xie, Zeshun Zong, Michal Yarom, Yonatan Bitton, Chenfanfu Jiang, Yizhou Sun, Kai-Wei Chang, and Aditya Grover.
\newblock Videophy: Evaluating physical commonsense for video generation.
\newblock In \emph{ICRL}, pages 102075--102121, 2025{\natexlab{a}}.

\bibitem[Bansal et~al.(2025{\natexlab{b}})Bansal, Peng, Bitton, Goldenberg, Grover, and Chang]{Bansal2025videophy2}
Hritik Bansal, Clark Peng, Yonatan Bitton, Roman Goldenberg, Aditya Grover, and Kai-Wei Chang.
\newblock {VideoPhy-2: A Challenging Action-Centric Physical Commonsense Evaluation in Video Generation}.
\newblock \emph{arXiv preprint arXiv:2503.06800}, 2025{\natexlab{b}}.

\bibitem[Cao et~al.(2018)Cao, Shen, Xie, Parkhi, and Zisserman]{Cao2018VGGFace2}
Qiong Cao, Li Shen, Weidi Xie, Omkar~M. Parkhi, and Andrew Zisserman.
\newblock Vggface2: A dataset for recognising faces across pose and age.
\newblock In \emph{FG}, 2018.

\bibitem[Caron et~al.(2021)Caron, Touvron, Misra, J\'egou, Mairal, Bojanowski, and Joulin]{Caron2021Dino}
Mathilde Caron, Hugo Touvron, Ishan Misra, Herv\'e J\'egou, Julien Mairal, Piotr Bojanowski, and Armand Joulin.
\newblock {Emerging Properties in Self-Supervised Vision Transformers}.
\newblock In \emph{ICCV}, 2021.

\bibitem[Chen et~al.(2024)Chen, Huang, Liu, Shen, Zhao, and Zhao]{Chen2024AnyDoor}
Xi Chen, Lianghua Huang, Yu Liu, Yujun Shen, Deli Zhao, and Hengshuang Zhao.
\newblock {AnyDoor: Zero-shot Object-level Image Customization}.
\newblock In \emph{CVPR}, 2024.

\bibitem[Cho et~al.(2023)Cho, Zala, and Bansal]{cho2023DALL-EVAL}
Jaemin Cho, Abhay Zala, and Mohit Bansal.
\newblock {DALL-EVAL: Probing the Reasoning Skills and Social Biases of Text-to-Image Generation Models}.
\newblock In \emph{ICCV}, 2023.

\bibitem[Deng et~al.(2022)Deng, Guo, Yang, Xue, Kotsia, and Zafeiriou]{Dengw2022arcface}
Jiankang Deng, Jia Guo, Jing Yang, Niannan Xue, Irene Kotsia, and Stefanos Zafeiriou.
\newblock {ArcFace: Additive Angular Margin Loss for Deep Face Recognition}.
\newblock \emph{IEEE TPAMI}, 44\penalty0 (10\_{Part\_1}):\penalty0 5962–5979, 2022.

\bibitem[Gokhale et~al.(2022)Gokhale, Palangi, Nushi, Vineet, Horvitz, Kamar, Baral, and Yang]{gokhale2022benchmarking}
Tejas Gokhale, Hamid Palangi, Besmira Nushi, Vibhav Vineet, Eric Horvitz, Ece Kamar, Chitta Baral, and Yezhou Yang.
\newblock Benchmarking spatial relationships in text-to-image generation.
\newblock \emph{arXiv preprint arXiv:2212.10015}, 2022.

\bibitem[Gu et~al.(2020)Gu, Bao, Chen, and Wen]{Gu2020QS}
Shuyang Gu, Jianmin Bao, Dong Chen, and Fang Wen.
\newblock {GIQA: Generated Image Quality Assessment}.
\newblock In \emph{ECCV}, 2020.

\bibitem[Gu et~al.(2023)Gu, Wang, Wu, Shi, Chen, Fan, Xiao, Zhao, Chang, Wu, Ge, Shan, and Shou]{Gu2023Mix-of-show}
Yuchao Gu, Xintao Wang, Jay~Zhangjie Wu, Yujun Shi, Yunpeng Chen, Zihan Fan, Wuyou Xiao, Rui Zhao, Shuning Chang, Weijia Wu, Yixiao Ge, Ying Shan, and Mike~Zheng Shou.
\newblock {Mix-of-show: decentralized low-rank adaptation for multi-concept customization of diffusion models}.
\newblock In \emph{NeurIPS}, 2023.

\bibitem[He et~al.(2024{\natexlab{a}})He, Geng, and Bo]{he2024uniportrait}
Junjie He, Yifeng Geng, and Liefeng Bo.
\newblock {UniPortrait: A Unified Framework for Identity-Preserving Single- and Multi-Human Image Personalization}.
\newblock \emph{arXiv preprint arXiv:2408.05939}, 2024{\natexlab{a}}.

\bibitem[He et~al.(2024{\natexlab{b}})He, Jiang, Zhang, Ku, Soni, Siu, Chen, Chandra, Jiang, Arulraj, Wang, Do, Ni, Lyu, Narsupalli, Fan, Lyu, Lin, and Chen]{He2024VideoScore}
Xuan He, Dongfu Jiang, Ge Zhang, Max Ku, Achint Soni, Sherman Siu, Haonan Chen, Abhranil Chandra, Ziyan Jiang, Aaran Arulraj, Kai Wang, Quy~Duc Do, Yuansheng Ni, Bohan Lyu, Yaswanth Narsupalli, Rongqi Fan, Zhiheng Lyu, Bill~Yuchen Lin, and Wenhu Chen.
\newblock {VideoScore: Building Automatic Metrics to Simulate Fine-grained Human Feedback for Video Generation}.
\newblock In \emph{EMNLP}, 2024{\natexlab{b}}.

\bibitem[Hinz et~al.(2022)Hinz, Heinrich, and Wermter]{Hinz2022SOAIscore}
Tobias Hinz, Stefan Heinrich, and Stefan Wermter.
\newblock {Semantic Object Accuracy for Generative Text-to-Image Synthesis}.
\newblock \emph{IEEE TPAMI}, 44\penalty0 (3):\penalty0 1552--1565, 2022.

\bibitem[Ho et~al.(2020)Ho, Jain, and Abbeel]{Ho2020Denoising}
Jonathan Ho, Ajay Jain, and Pieter Abbeel.
\newblock {Denoising Diffusion Probabilistic Models}.
\newblock In \emph{NeurIPS}, 2020.

\bibitem[Hu et~al.(2023)Hu, Liu, Kasai, Wang, Ostendorf, Krishna, and Smith]{Hu2023TIFA}
Yushi Hu, Benlin Liu, Jungo Kasai, Yizhong Wang, Mari Ostendorf, Ranjay Krishna, and Noah~A. Smith.
\newblock {TIFA: Accurate and Interpretable Text-to-Image Faithfulness Evaluation with Question Answering}.
\newblock In \emph{ICCV}, 2023.

\bibitem[Karras et~al.(2019)Karras, Laine, and Aila]{Karras2019Flickr-Faces-HQ}
Tero Karras, Samuli Laine, and Timo Aila.
\newblock {A Style-Based Generator Architecture for Generative Adversarial Networks}.
\newblock In \emph{CVPR}, 2019.

\bibitem[Kirillov et~al.(2023)Kirillov, Mintun, Ravi, Mao, Rolland, Gustafson, Xiao, Whitehead, Berg, Lo, Dollar, and Girshick]{Kirillov2023sam}
Alexander Kirillov, Eric Mintun, Nikhila Ravi, Hanzi Mao, Chloe Rolland, Laura Gustafson, Tete Xiao, Spencer Whitehead, Alexander~C. Berg, Wan-Yen Lo, Piotr Dollar, and Ross Girshick.
\newblock {Segment Anything}.
\newblock In \emph{ICCV}, 2023.

\bibitem[Kong et~al.(2024)Kong, Zhang, Yang, Wang, Zhang, Wu, Chen, Liu, and Luo]{Kong2024OMG}
Zhe Kong, Yong Zhang, Tianyu Yang, Tao Wang, Kaihao Zhang, Bizhu Wu, Guanying Chen, Wei Liu, and Wenhan Luo.
\newblock {OMG: Occlusion-Friendly Personalized Multi-concept Generation in Diffusion Models}.
\newblock In \emph{ECCV}, 2024.

\bibitem[Ku et~al.(2024{\natexlab{a}})Ku, Jiang, Wei, Yue, and Chen]{ku2024viescore}
Max Ku, Dongfu Jiang, Cong Wei, Xiang Yue, and Wenhu Chen.
\newblock {VIES}core: Towards explainable metrics for conditional image synthesis evaluation.
\newblock In \emph{ACL}, pages 12268--12290. ACL, 2024{\natexlab{a}}.

\bibitem[Ku et~al.(2024{\natexlab{b}})Ku, Li, Zhang, Lu, Fu, Zhuang, and Chen]{ku2024imagenhub}
Max Ku, Tianle Li, Kai Zhang, Yujie Lu, Xingyu Fu, Wenwen Zhuang, and Wenhu Chen.
\newblock Imagenhub: Standardizing the evaluation of conditional image generation models.
\newblock In \emph{ICLR}, 2024{\natexlab{b}}.

\bibitem[Kumari et~al.(2023)Kumari, Zhang, Zhang, Shechtman, and Zhu]{Kumari2023customdiffusion}
Nupur Kumari, Bingliang Zhang, Richard Zhang, Eli Shechtman, and Jun-Yan Zhu.
\newblock {Multi-Concept Customization of Text-to-Image Diffusion}.
\newblock In \emph{CVPR}, 2023.

\bibitem[Li et~al.(2023{\natexlab{a}})Li, Li, and Hoi]{Li2023BLIPdiffusion}
Dongxu Li, Junnan Li, and Steven~C.H. Hoi.
\newblock {BLIP-diffusion: pre-trained subject representation for controllable text-to-image generation and editing}.
\newblock In \emph{NeurIPS}, 2023{\natexlab{a}}.

\bibitem[Li et~al.(2023{\natexlab{b}})Li, Li, Savarese, and Hoi]{li2023blip2}
Junnan Li, Dongxu Li, Silvio Savarese, and Steven Hoi.
\newblock {{BLIP}-2: Bootstrapping Language-Image Pre-training with Frozen Image Encoders and Large Language Models}.
\newblock In \emph{ICML}, 2023{\natexlab{b}}.

\bibitem[Li et~al.(2021)Li, Wu, Koh, Tang, and Sun]{Li2021yoloscore}
Zejian Li, Jingyu Wu, Immanuel Koh, Yongchuan Tang, and Lingyun Sun.
\newblock {Image Synthesis From Layout With Locality-Aware Mask Adaption}.
\newblock In \emph{ICCV}, 2021.

\bibitem[Lin et~al.(2014)Lin, Maire, Belongie, Hays, Perona, Ramanan, Doll{\'a}r, and Zitnick]{LinCOCO2014}
Tsung-Yi Lin, Michael Maire, Serge Belongie, James Hays, Pietro Perona, Deva Ramanan, Piotr Doll{\'a}r, and C.~Lawrence Zitnick.
\newblock {Microsoft COCO: Common Objects in Context}.
\newblock In \emph{ECCV}, 2014.

\bibitem[Lin et~al.(2024)Lin, Pathak, Li, Li, Xia, Neubig, Zhang, and Ramanan]{Lin2024VQAScore}
Zhiqiu Lin, Deepak Pathak, Baiqi Li, Jiayao Li, Xide Xia, Graham Neubig, Pengchuan Zhang, and Deva Ramanan.
\newblock Evaluating text-to-visual generation with image-to-text generation.
\newblock In \emph{ECCV}, page 366–384, 2024.

\bibitem[Liu et~al.(2020)Liu, Shahroudy, Perez, Wang, Duan, and Kot]{Liu2020NTURGB+D120}
Jun Liu, Amir Shahroudy, Mauricio Perez, Gang Wang, Ling-Yu Duan, and Alex~C. Kot.
\newblock {NTU RGB+D 120: A Large-Scale Benchmark for 3D Human Activity Understanding}.
\newblock \emph{IEEE TPAMI}, 42\penalty0 (10):\penalty0 2684–2701, 2020.

\bibitem[Liu et~al.(2024)Liu, Zeng, Ren, Li, Zhang, Yang, Jiang, Li, Yang, Su, Zhu, and Zhang]{Liu2024GroundingDINO}
Shilong Liu, Zhaoyang Zeng, Tianhe Ren, Feng Li, Hao Zhang, Jie Yang, Qing Jiang, Chunyuan Li, Jianwei Yang, Hang Su, Jun Zhu, and Lei Zhang.
\newblock {Grounding dino: Marrying dino with grounded pre-training for open-set object detection}.
\newblock In \emph{ECCV}, 2024.

\bibitem[Liu et~al.(2023)Liu, Zhang, Shen, Zheng, Zhu, Feng, Liu, Zhao, Zhou, and Cao]{Liu2023Cones2}
Zhiheng Liu, Yifei Zhang, Yujun Shen, Kecheng Zheng, Kai Zhu, Ruili Feng, Yu Liu, Deli Zhao, Jingren Zhou, and Yang Cao.
\newblock {Cones 2: customizable image synthesis with multiple subjects}.
\newblock In \emph{NeurIPS}, 2023.

\bibitem[Ma et~al.(2024)Ma, Liang, Chen, and Lu]{Ma2024Subject-Diffusion}
Jian Ma, Junhao Liang, Chen Chen, and Haonan Lu.
\newblock {Subject-Diffusion: Open Domain Personalized Text-to-Image Generation without Test-time Fine-tuning}.
\newblock In \emph{SIGGRAPH}, 2024.

\bibitem[{Nema, Preksha and Khapra, Mitesh M.}(2018)]{Nema2018towards}
{Nema, Preksha and Khapra, Mitesh M.}
\newblock {Towards a Better Metric for Evaluating Question Generation Systems}.
\newblock In \emph{EMNLP}, 2018.

\bibitem[Ohi et~al.(2025)Ohi, Kaneko, Okazaki, and Inoue]{Ohi2024MultimodalMM}
Masanari Ohi, Masahiro Kaneko, Naoaki Okazaki, and Nakamasa Inoue.
\newblock {Multi-modal, Multi-task, Multi-criteria Automatic Evaluation with Vision Language Models}.
\newblock \emph{arXiv preprint arXiv:2412.14613}, 2025.

\bibitem[OpenAI et~al.(2024)OpenAI, Achiam, Adler, Agarwal, Ahmad, Akkaya, Aleman, Almeida, Altenschmidt, Altman, and et~al.]{openai2024gpt4technicalreport}
OpenAI, Josh Achiam, Steven Adler, Sandhini Agarwal, Lama Ahmad, Ilge Akkaya, Florencia~Leoni Aleman, Diogo Almeida, Janko Altenschmidt, Sam Altman, and Shyamal~Anadkat et al.
\newblock {GPT-4 Technical Report}.
\newblock \emph{arXiv preprint arXiv:2303.08774}, 2024.

\bibitem[Peng et~al.(2025)Peng, Cui, Tang, Qi, Dong, Bai, Han, Ge, Zhang, and Xia]{peng2024dreambench}
Yuang Peng, Yuxin Cui, Haomiao Tang, Zekun Qi, Runpei Dong, Jing Bai, Chunrui Han, Zheng Ge, Xiangyu Zhang, and Shu-Tao Xia.
\newblock Dreambench++: A human-aligned benchmark for personalized image generation.
\newblock In \emph{ICLR}, 2025.

\bibitem[Phung et~al.(2024)Phung, Ge, and Huang]{Phung2024AttentionRefocusing}
Quynh Phung, Songwei Ge, and Jia-Bin Huang.
\newblock {Grounded Text-to-Image Synthesis with Attention Refocusing}.
\newblock In \emph{CVPR}, 2024.

\bibitem[Plummer et~al.(2015)Plummer, Wang, Cervantes, Caicedo, Hockenmaier, and Lazebnik]{Plummer2015Flickr30k}
Bryan~A. Plummer, Liwei Wang, Chris~M. Cervantes, Juan~C. Caicedo, Julia Hockenmaier, and Svetlana Lazebnik.
\newblock {Flickr30k Entities: Collecting Region-to-Phrase Correspondences for Richer Image-to-Sentence Models}.
\newblock In \emph{ICCV}, 2015.

\bibitem[Podell et~al.(2023)Podell, English, Lacey, Blattmann, Dockhorn, M{\"u}ller, Penna, and Rombach]{podell2023sdxl}
Dustin Podell, Zion English, Kyle Lacey, Andreas Blattmann, Tim Dockhorn, Jonas M{\"u}ller, Joe Penna, and Robin Rombach.
\newblock {Sdxl: Improving latent diffusion models for high-resolution image synthesis}.
\newblock \emph{arXiv preprint arXiv:2307.01952}, 2023.

\bibitem[Radford et~al.(2021)Radford, Kim, Hallacy, Ramesh, Goh, Agarwal, Sastry, Askell, Mishkin, Clark, Krueger, and Sutskever]{Radford2021CLIP}
Alec Radford, Jong~Wook Kim, Chris Hallacy, Aditya Ramesh, Gabriel Goh, Sandhini Agarwal, Girish Sastry, Amanda Askell, Pamela Mishkin, Jack Clark, Gretchen Krueger, and Ilya Sutskever.
\newblock {Learning Transferable Visual Models From Natural Language Supervision}.
\newblock In \emph{ICML}, 2021.

\bibitem[Rombach et~al.(2022)Rombach, Blattmann, Lorenz, Esser, and Ommer]{rombach2022highresolution}
Robin Rombach, Andreas Blattmann, Dominik Lorenz, Patrick Esser, and Bj\"orn Ommer.
\newblock {High-Resolution Image Synthesis With Latent Diffusion Models}.
\newblock In \emph{CVPR}, 2022.

\bibitem[Ruiz et~al.(2023)Ruiz, Li, Jampani, Pritch, Rubinstein, and Aberman]{ruiz2023dreambooth}
Nataniel Ruiz, Yuanzhen Li, Varun Jampani, Yael Pritch, Michael Rubinstein, and Kfir Aberman.
\newblock {Dreambooth: Fine tuning text-to-image diffusion models for subject-driven generation}.
\newblock In \emph{CVPR}, 2023.

\bibitem[Saharia et~al.(2022)Saharia, Chan, Saxena, Lit, Whang, Denton, Ghasemipour, Ayan, Mahdavi, Gontijo-Lopes, Salimans, Ho, Fleet, and Norouzi]{saharia2022photorealistic}
Chitwan Saharia, William Chan, Saurabh Saxena, Lala Lit, Jay Whang, Emily Denton, Seyed Kamyar~Seyed Ghasemipour, Burcu~Karagol Ayan, S.~Sara Mahdavi, Raphael Gontijo-Lopes, Tim Salimans, Jonathan Ho, David~J Fleet, and Mohammad Norouzi.
\newblock {Photorealistic text-to-image diffusion models with deep language understanding}.
\newblock In \emph{NeurIPS}, 2022.

\bibitem[Schuhmann et~al.(2021)Schuhmann, Vencu, Beaumont, Kaczmarczyk, Mullis, Katta, Coombes, Jitsev, and Komatsuzaki]{Schuhmann2021LAION400MOD}
Christoph Schuhmann, Richard Vencu, Romain Beaumont, Robert Kaczmarczyk, Clayton Mullis, Aarush Katta, Theo Coombes, Jenia Jitsev, and Aran Komatsuzaki.
\newblock {LAION-400M: Open Dataset of CLIP-Filtered 400 Million Image-Text Pairs}.
\newblock \emph{arXiv preprint arXiv:2111.02114}, 2021.

\bibitem[Schuhmann et~al.(2022)Schuhmann, Beaumont, Vencu, Gordon, Wightman, Cherti, Coombes, Katta, Mullis, Wortsman, Schramowski, Kundurthy, Crowson, Schmidt, Kaczmarczyk, and Jitsev]{schuhmann2022laion5bopenlargescaledataset}
Christoph Schuhmann, Romain Beaumont, Richard Vencu, Cade Gordon, Ross Wightman, Mehdi Cherti, Theo Coombes, Aarush Katta, Clayton Mullis, Mitchell Wortsman, Patrick Schramowski, Srivatsa Kundurthy, Katherine Crowson, Ludwig Schmidt, Robert Kaczmarczyk, and Jenia Jitsev.
\newblock {LAION-5B: An open large-scale dataset for training next generation image-text models}.
\newblock \emph{arXiv preprint arXiv:2210.08402}, 2022.

\bibitem[Shahroudy et~al.(2016)Shahroudy, Liu, Ng, and Wang]{Shahroudy2016NTURGB}
Amir Shahroudy, Jun Liu, Tian-Tsong Ng, and Gang Wang.
\newblock {NTU RGB+D: A Large Scale Dataset for 3D Human Activity Analysis}.
\newblock In \emph{CVPR}, 2016.

\bibitem[Sohl-Dickstein et~al.(2015)Sohl-Dickstein, Weiss, Maheswaranathan, and Ganguli]{Sohl-Dickstein2015Deep}
Jascha Sohl-Dickstein, Eric~A. Weiss, Niru Maheswaranathan, and Surya Ganguli.
\newblock {Deep unsupervised learning using nonequilibrium thermodynamics}.
\newblock In \emph{ICML}, 2015.

\bibitem[Szegedy et~al.(2014)Szegedy, Liu, Jia, Sermanet, Reed, Anguelov, Erhan, Vanhoucke, and Rabinovich]{szegedy2014goingdeeperconvolutions}
Christian Szegedy, Wei Liu, Yangqing Jia, Pierre Sermanet, Scott Reed, Dragomir Anguelov, Dumitru Erhan, Vincent Vanhoucke, and Andrew Rabinovich.
\newblock {Going Deeper with Convolutions}.
\newblock \emph{arXiv preprint arXiv:1409.4842}, 2014.

\bibitem[Team et~al.(2025)Team, Anil, Borgeaud, Alayrac, Yu, Soricut, Schalkwyk, Dai, Hauth, Millican, and et~al.]{geminiteam2025geminifamilyhighlycapable}
Gemini Team, Rohan Anil, Sebastian Borgeaud, Jean-Baptiste Alayrac, Jiahui Yu, Radu Soricut, Johan Schalkwyk, Andrew~M. Dai, Anja Hauth, Katie Millican, and David~Silver et al.
\newblock Gemini: A family of highly capable multimodal models.
\newblock \emph{arXiv preprint arXiv:2312.11805}, 2025.

\bibitem[von Platen et~al.(2022)von Platen, Patil, Lozhkov, Cuenca, Lambert, Rasul, Davaadorj, Nair, Paul, Berman, Xu, Liu, and Wolf]{platen2022diffusers}
Patrick von Platen, Suraj Patil, Anton Lozhkov, Pedro Cuenca, Nathan Lambert, Kashif Rasul, Mishig Davaadorj, Dhruv Nair, Sayak Paul, William Berman, Yiyi Xu, Steven Liu, and Thomas Wolf.
\newblock {Diffusers: State-of-the-art diffusion models}.
\newblock \emph{GitHub repository}, 2022.

\bibitem[Voynov et~al.(2023)Voynov, Chu, Cohen-Or, and Aberman]{voynov2023p+}
Andrey Voynov, Qinghao Chu, Daniel Cohen-Or, and Kfir Aberman.
\newblock {P+: Extended Textual Conditioning in Text-to-Image Generation}.
\newblock \emph{arXiv preprint arXiv:2303.09522}, 2023.

\bibitem[Wang et~al.(2024)Wang, Duan, Zhai, Wang, and Min]{Wang2024AIGV-Assessor}
Jiarui Wang, Huiyu Duan, Guangtao Zhai, Juntong Wang, and Xiongkuo Min.
\newblock {AIGV-Assessor: Benchmarking and Evaluating the Perceptual Quality of Text-to-Video Generation with LMM}.
\newblock \emph{arXiv preprint arXiv:2411.17221}, 2024.

\bibitem[Wang et~al.(2025)Wang, Yang, Wang, Xu, Wang, Wang, Luo, Zhang, Hu, and Zhang]{wang2025cigeval}
Jifang Wang, Xue Yang, Longyue Wang, Zhenran Xu, Yiyu Wang, Yaowei Wang, Weihua Luo, Kaifu Zhang, Baotian Hu, and Min Zhang.
\newblock A unified agentic framework for evaluating conditional image generation.
\newblock \emph{arXiv preprint arXiv:2504.07046}, 2025.

\bibitem[Winata et~al.(2025)Winata, Anugraha, Susanto, Kuwanto, and Wijaya]{winata2025metametrics}
Genta~Indra Winata, David Anugraha, Lucky Susanto, Garry Kuwanto, and Derry~Tanti Wijaya.
\newblock {MetaMetrics: Calibrating Metrics For Generation Tasks Using Human Preferences}.
\newblock \emph{arXiv preprint arXiv:2410.02381}, 2025.

\bibitem[Wu et~al.(2024)Wu, Zhang, Zhang, Chen, Liao, Li, Gao, Wang, Zhang, Sun, Yan, Min, Zhai, and Lin]{wu2024qalign}
Haoning Wu, Zicheng Zhang, Weixia Zhang, Chaofeng Chen, Liang Liao, Chunyi Li, Yixuan Gao, Annan Wang, Erli Zhang, Wenxiu Sun, Qiong Yan, Xiongkuo Min, Guangtao Zhai, and Weisi Lin.
\newblock {Q-Align: Teaching {LMM}s for Visual Scoring via Discrete Text-Defined Levels}.
\newblock In \emph{ICML}, 2024.

\bibitem[Xiao et~al.(2024)Xiao, Yin, Freeman, Durand, and Han]{Xiao2024FastComposer}
Guangxuan Xiao, Tianwei Yin, William~T. Freeman, Fr\'{e}do Durand, and Song Han.
\newblock {FastComposer: Tuning-Free Multi-subject Image Generation with Localized Attention}.
\newblock \emph{IJCV}, 133\penalty0 (3):\penalty0 1175–1194, 2024.

\bibitem[Xu et~al.(2023)Xu, Liu, Wu, Tong, Li, Ding, Tang, and Dong]{xu2023imagereward}
Jiazheng Xu, Xiao Liu, Yuchen Wu, Yuxuan Tong, Qinkai Li, Ming Ding, Jie Tang, and Yuxiao Dong.
\newblock Imagereward: learning and evaluating human preferences for text-to-image generation.
\newblock In \emph{NeurIPS}, pages 15903--15935, 2023.

\bibitem[Young et~al.(2014)Young, Lai, Hodosh, and Hockenmaier]{young2014From}
Peter Young, Alice Lai, Micah Hodosh, and Julia Hockenmaier.
\newblock From image descriptions to visual denotations: New similarity metrics for semantic inference over event descriptions.
\newblock \emph{TACL}, 2:\penalty0 67--78, 2014.

\bibitem[Yu(2025)]{yu2025llmsreallythinkstepbystep}
Yijiong Yu.
\newblock Do llms really think step-by-step in implicit reasoning?
\newblock \emph{arXiv preprint arXiv:2411.15862}, 2025.

\bibitem[Zhang et~al.(2016)Zhang, Zhang, Li, and Qiao]{Zhang2016mtcnn}
Kaipeng Zhang, Zhanpeng Zhang, Zhifeng Li, and Yu Qiao.
\newblock Joint face detection and alignment using multitask cascaded convolutional networks.
\newblock \emph{SPL}, 23:\penalty0 1499--1503, 2016.

\bibitem[Zhang et~al.(2023)Zhang, Rao, and Agrawala]{Zhang2023ControleNet}
Lvmin Zhang, Anyi Rao, and Maneesh Agrawala.
\newblock {Adding Conditional Control to Text-to-Image Diffusion Models}.
\newblock In \emph{ICCV}, 2023.

\bibitem[Zheng et~al.(2023)Zheng, Chiang, Sheng, Zhuang, Wu, Zhuang, Lin, Li, Li, Xing, Zhang, Gonzalez, and Stoica]{Lianmin2023NeurIPS}
Lianmin Zheng, Wei-Lin Chiang, Ying Sheng, Siyuan Zhuang, Zhanghao Wu, Yonghao Zhuang, Zi Lin, Zhuohan Li, Dacheng Li, Eric Xing, Hao Zhang, Joseph~E Gonzalez, and Ion Stoica.
\newblock {Judging LLM-as-a-Judge with MT-Bench and Chatbot Arena}.
\newblock In \emph{NeurIPS}, 2023.

\bibitem[Zhong et~al.(2024)Zhong, Shen, Wang, Lu, Jiao, Ouyang, Yu, Han, and Chen]{zhong2024multilora}
Ming Zhong, Yelong Shen, Shuohang Wang, Yadong Lu, Yizhu Jiao, Siru Ouyang, Donghan Yu, Jiawei Han, and Weizhu Chen.
\newblock {Multi-LoRA Composition for Image Generation}.
\newblock \emph{arXiv preprint arXiv:2402.16843}, 2024.

\bibitem[Zhou et~al.(2022)Zhou, Koltun, and Kr{\"a}henb{\"u}hl]{zhou2021simple}
Xingyi Zhou, Vladlen Koltun, and Philipp Kr{\"a}henb{\"u}hl.
\newblock {Simple multi-dataset detection}.
\newblock In \emph{CVPR}, 2022.

\end{thebibliography}
